\documentclass{article}

    \PassOptionsToPackage{numbers, compress}{natbib}
    \usepackage[preprint]{neurips_2022}

\usepackage[utf8]{inputenc} % allow utf-8 input
\usepackage[T1]{fontenc}    % use 8-bit T1 fonts
\usepackage{hyperref}       % hyperlinks
\usepackage{url}            % simple URL typesetting
\usepackage{booktabs}       % professional-quality tables
\usepackage{amsfonts}       % blackboard math symbols
\usepackage{nicefrac}       % compact symbols for 1/2, etc.
\usepackage{microtype}      % microtypography
\usepackage{xcolor}         % colors
\usepackage{subcaption} %% For complex figures with subfigures/subcaptions
\usepackage{listings}
\usepackage{relsize}
\usepackage{enumerate}
\usepackage{float}
\usepackage{graphicx}
\usepackage{multirow}
\usepackage[algosection,ruled,vlined,linesnumbered,algo2e]{algorithm2e}
  \SetKwInput{KwIn}{Inputs}
  \SetKwFunction{Fthreat}{ThreatModel}
  \SetKwFunction{Feval}{Eval}
  \SetKwFunction{Ffor}{for}
  \SetKwProg{Fn}{}{:}{}
  \DontPrintSemicolon
\usepackage{wrapfig}
\usepackage{amsmath}
\usepackage{amssymb}
\usepackage{mathtools}
\usepackage{amsthm}
\usepackage{dsfont}
\usepackage{wasysym}
\usepackage{cleveref}

\theoremstyle{plain}
\newtheorem{theorem}{Theorem}[section]
\newtheorem{proposition}[theorem]{Proposition}

\theoremstyle{definition}
\newtheorem{definition}[theorem]{Definition}

\theoremstyle{remark}

\title{Faithful Explanations for Deep Graph Models}

\author{%
  Zifan Wang, Yuhang Yao, Chaoran Zhang, Han Zhang, Youjie Kang\\
  \textbf{Carlee Joe-Wong}, \textbf{Matt Fredrikson}, \textbf{Anupam Datta}\\
  Electrical and Computer Engineering\\
  Carnegie Mellon University\\
  Pittsburgh, PA 15213 \\
  \texttt{zifan@cmu.edu}\\
}

\begin{document}

\maketitle

\begin{abstract}
This paper studies faithful explanations for Graph Neural Networks (GNNs). First, we provide a new and general method for formally characterizing the faithfulness of explanations for GNNs. It applies to existing explanation methods, including feature attributions and subgraph explanations. Second, our analytical and empirical results demonstrate that feature attribution methods cannot capture the nonlinear effect of edge features, while existing subgraph explanation methods are not faithful. Third, we introduce \emph{k-hop Explanation with a Convolutional Core} (KEC), a new explanation method that provably maximizes faithfulness to the original GNN by leveraging information about the graph structure in its adjacency matrix and its \emph{k-th} power. Lastly, our empirical results over both synthetic and real-world datasets for classification and anomaly detection tasks with GNNs demonstrate the effectiveness of our approach.
\end{abstract}
\section{Introduction}
Graph Neural Networks (GNNs) have shown promise in a wide range of prediction tasks, ranging from applications in social networking~\citep{fan2019graph, Mu2019GraphAN, Huang2021KnowledgeawareCG} to cyber-physical systems security~\citep{9462385, yao2022fedgcn}.
However, their predictions remain opaque to human users, which threatens to ultimately limit their successful adoption in many domains.
A widely-used approach for explaining the predictions of neural networks computes \emph{feature attributions}~\citep{simonyan2013deep, leino2018influence, sundararajan2017axiomatic, smilkov2017smoothgrad, Gradcam, lime, NIPS2017_7062, wang2020score, Petsiuk2018RISERI, Fong2017InterpretableEO}, which are interpreted as importance scores for the input features relative to a given quantity of interest, such as the logit of a predicted label. In the context of GNNs, recent work has proposed numerous \emph{subgraph explanation} techniques~\citep{ying2019gnnexplainer, luo2020parameterized, PGMExplainer, alsentzer2020subgraph}, which aim to specifically leverage graph structure when identifying the node and edge features that are important for a prediction.

% For any explanation method to be applied to make sense of predictions made by GNNs, it must be faithful to begin with. Namely, a method is faithful if the information that it provides is consistent with the actual behavior of the graph model.

For any explanation method to be useful in diagnosing and improving the graph models' performance, it must be \emph{faithful} to begin with. Namely, the information provided by an explanation should be consistent with the behavior of the underlying model. Metrics related to faithfulness have been widely researched for attribution methods used in vision models~\citep{Wang2020InterpretingIO, leino2018influence, sundararajan2017axiomatic,NEURIPS2019_a7471fdc, ancona2018towards, Hooker2019ABF} but have received limited attention for graph models. 
Thus, the question that we aim to address in this paper is, \emph{how should one construct and evaluate faithful explanations for GNNs?}

\begin{figure}[t]
    \centering
    \begin{subfigure}[b]{0.4\textwidth}
        \centering
        \includegraphics[width=\textwidth]{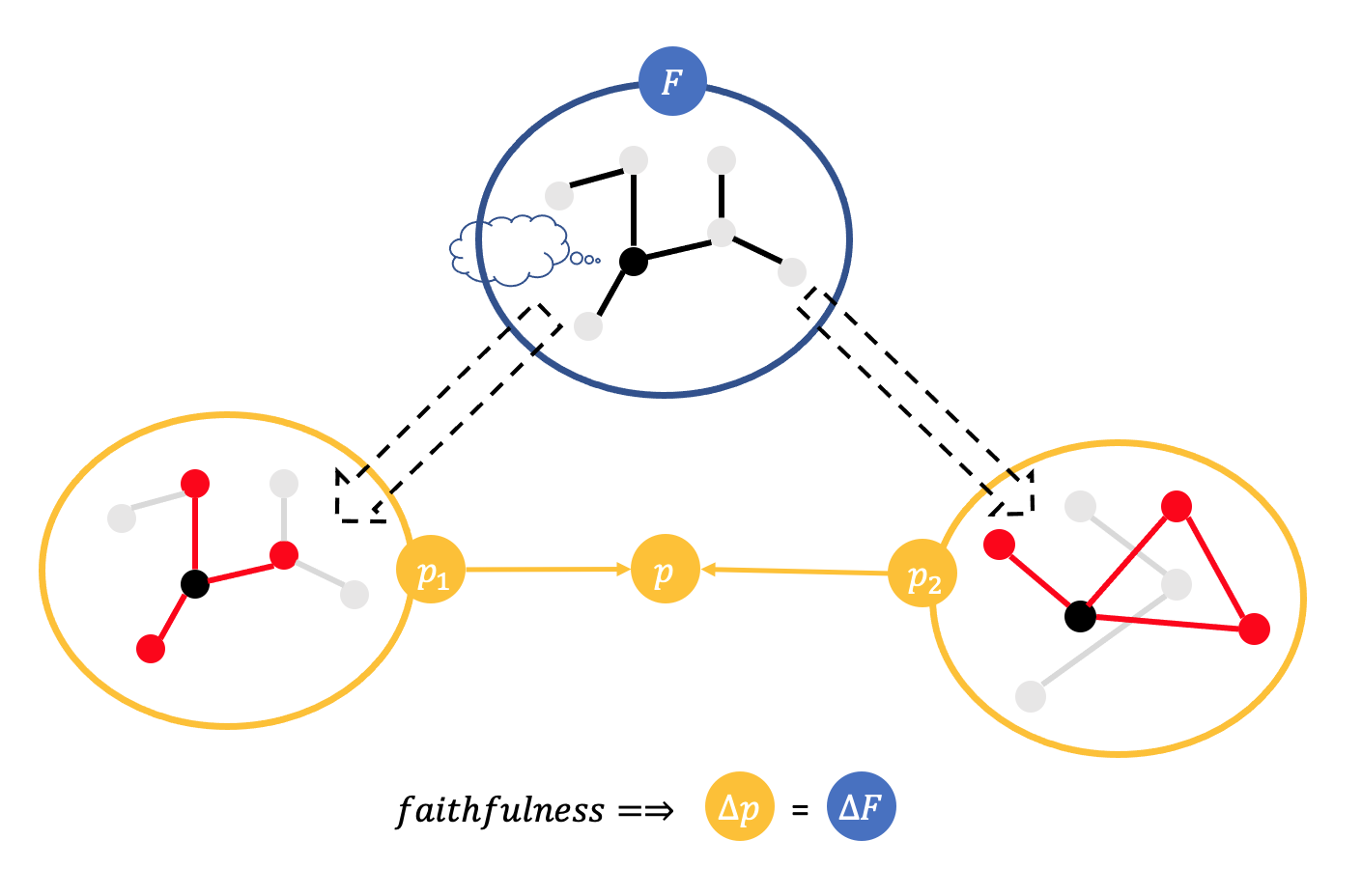}\\
        \subcaption{}
    \end{subfigure}%
    \begin{subfigure}[b]{0.5\textwidth}
        \centering
        \includegraphics[width=\textwidth]{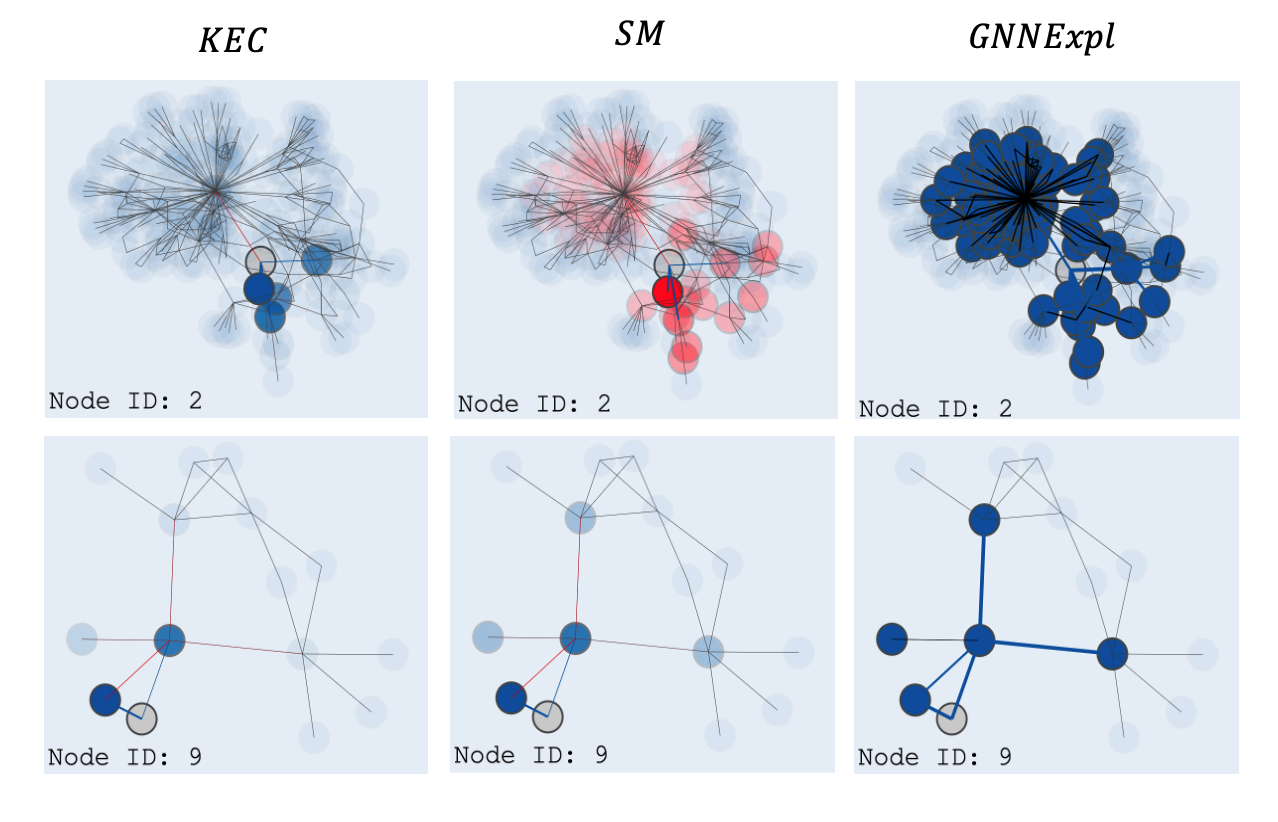}\\
        \subcaption{}
    \end{subfigure}%
    \caption{(a) An illustration of the proposed faithful graph explanation KEC. We consider a target graph model $F$ where messages pass as far as from a node's 2-hop neighbors, and an quantity of interest, the prediction on the black node. KEC firstly builds two functions $p_1$ and $p_2$ which directly take the information from its 1-hop and 2-hop neighbors, respectively. Secondly KEC minimizes the difference between the change $\Delta p = \Delta p_1 + \Delta p_2$ and the change $\Delta F$ under the local perturbations to the input graph. Finally, the optimal parameters of $p_1$ and $p_2$ that minimizes $(\Delta p -\Delta F)^2$ are used to compute the importance of node and edge features in the given input. (b) an visualization of the proposed method KEC compared with Saliency Map(SM) and GNNExpl~\cite{ying2019gnnexplainer} on Cora. Visualizations of other methods are not included for the space limit in this version. Node to explain is colored in gray. Nodes and edges with positive influence are in blue and the negative ones are in red. The opacity of the nodes and the thickness of edges correspond to actual influence scores.}
    \label{fig:illustration}
\end{figure}

To answer this question, we begin by introducing a unified faithfulness metric for both gradient-based and subgraph-based attribution methods (Sec.~\ref{sec:general-faithfulness}). We then look towards applying this metric to the task of constructing explanations, by treating it as an objective function to optimize. A key challenge in this agenda stems from the fact that GNNs are locally linear w.r.t node features, but in general this is not the case w.r.t their edge features (Prop.~\ref{prop:non-linearity-of-A}). This finding motivates a focus on explaining GNNs via non-linear local approximations, as these can more faithfully capture the importance of edge features in GNN predictions.

Our contributions are summarized as follows: \emph{1)} We introduce \emph{general faithfulness}, a quantitative metric for both subgraph and gradient-based GNN explanations; \emph{2)} We introduce a novel GNN explanation technique called \emph{\textbf{K}-hop \textbf{E}xplanation with a \textbf{C}onvolutional Core} (KEC) that maximizes this metric; \emph{3)} We evaluate KEC on synthetic and real-world datasets showing that KEC is up to 3 OOM more faithful than prior work (Table.~\ref{tab:gf-individual}). Notably, we find that some existing methods known for faithfulness when applied to vision models, e.g. Integrated Gradients~\cite{sundararajan2017axiomatic}, are not as faithful on GNNs. 
Additionally, We find that KEC is often significantly more faithful than subgraph explanations.
Finally, we provide a case study on anomaly detection models in an industrial control setting, and highlight the insights that our approach can provide in understanding false positive predictions (Sec.~\ref{sec: case-study}).

\section{Background}\label{sec:background}

We provide background on the notation used throughout the paper, and then outline the predominant explanation techniques given by prior work both for general feed-forward networks and GNNs.
Finally, we discuss notions of explanation faithfulness used in prior work on feature attribution.

\paragraph{Notation.} Let $(X\in \mathbb{R}^{n\times d}, \text A \in \mathbb{R}^{n\times n}_{+})$ be a graph of $n$ nodes and $d$ features, with adjacency matrix $\text A$. 
Note that we do not require adjacency information to be discrete, as several GNN frameworks (e.g., Pytorch Geometric~\citep{Fey/Lenssen/2019}), treat edges as positive scalars.
In this paper, we are primarily interested in the node classification task. 
Given a graph $(X, \text A)$, the model aims to predict a nominal-valued label for each node. 
Formally, we consider a Graph Convolution Network $y = \arg\max_c \{e^\top_vf(X, A)\}_c$ where $f$ is a stack of graph convolution and dense layers, separated by ReLU activations, and $e^\top_v$ selects the prediction of the target node. 
For instance, when $e_v[i] = \mathbb{I}[i=k]$, $y$ is the prediction of the $k$-th node. When $e_v$ is clear from the context and the class of interest is the network's prediction, we write $F(X, A)= \{e^\top_vf(X, A)\}_y$.
The graph convolution that we assume is expressed as $N(A)XW$, where $N(A) = D^{-1/2}(A+I)D^{-1/2}$, $D$ is the degree matrix, and $W$ is the kernel~\cite{kipf2017semi}.

We write $\mathbb{V}(H)$ to denote the vectorization of a matrix $H$ by stacking its column vectors, and $\mathbb{V}^{-1}$ for the corresponding inverse.
Finally, given a distribution $\mathcal{D}$ with support in $\mathbb{R}^d$,
we write $B(x, \mathcal{D})$ to denote the neighborhood $\{x'|x'=x- \epsilon, \epsilon \sim \mathcal{D}\}$ around $x$.

Throughout the paper we consider the following explanation methods for graph networks.

\paragraph{Feature Attributions.} The output of \emph{feature attribution} is a score for each feature that indicates its importance for increasing a quantity of interest, e.g. the output of prediction score. Among the space of feature attributions we consider \emph{distributional influence}~\citep{leino2018influence} because they are axiomatically justified for some desired properties, e.g. \emph{linear agreement}~\citep{leino2018influence}, and their faithfulness to the target function is provable (discussions to follow in Sec.~\ref{sec:background-attribution-faithfulness}). We adapt \emph{distributional influence} for graph models in Def.~\ref{def:distributional-influence}.

\begin{definition}[Distributional Influence~\citep{leino2018influence}]\label{def:distributional-influence}
Given a graph $(X, \text A)$, let $B(X, \mathcal{D}_X)$ and $B(A, \mathcal{D}_A)$ be neighborhoods around its node features and edge weights, respectively. 
Its \emph{feature influence} $\chi(X, \mathcal{D}_X)$ and \emph{edge influence} $\chi(A, \mathcal{D}_A)$ for a the model $F(X, A)$ are defined as $\chi(X, \mathcal{D}_X) = \mathbb{E}_{\epsilon \sim \mathcal{D}_X} \triangledown_{X} F(X+\epsilon, A)$ and $\chi(A, \mathcal{D}_A) = \mathbb{E}_{\epsilon \sim \mathcal{D}_A} \triangledown_{A} F(X, A+\epsilon)$.

% $\chi(X, \mathcal{D}_X) = \mathbb{E}_{\epsilon \sim \mathcal{D}_X} \triangledown_{X} F(X+\epsilon, A), 
% \chi(A, \mathcal{D}_A)= \mathbb{E}_{\epsilon \sim \mathcal{D}_A} \triangledown_{A} F(X, A+\epsilon)
% $.
\end{definition}

Special cases of points defined by the distribution of interest derives leads to a set of existing methods and we are interested in the following two in the rest of the paper: 1) a set only containing the input of interest -- Saliency Map (SM)~\citep{simonyan2013deep}; and 2) a uniform distribution over a linear path from a baseline point to the input -- the integral part of Integrated Gradient (IG)~\citep{sundararajan2017axiomatic}. We discuss our choices of the baselines for IG in Sec.~\ref{sec:evaluations}.

\paragraph{Subgraph Explanations.} The output of a subgraph explanation is a perturbed matrix of node features or/and an adjacency matrix with more zero entries, which are considered as the most relevant part of the input graph towards the output~\citep{ying2019gnnexplainer, luo2020parameterized, PGMExplainer, alsentzer2020subgraph}. One way to unify several subgraph explanations is shown in Def.~\ref{def:subgraph-explanation}.

\begin{definition}[Subgraph Explanation]\label{def:subgraph-explanation}
Given the output of graph network $F$ and an input $(X, A)$, the subgraph graph explanation is an instance $(X_s, A_s)$ that minimize the following objective:
\begin{align}
    \min_{X_s, A_s} L(X_s, A_s, F)\quad \quad s.t. \quad s(X_s, A_s) \leq K
\end{align}
where $s$ and $L$ are a user-defined size function and a performance penalty, respectively. $K$ is a constant.
\end{definition}

The size function $s(\cdot, \cdot)$ ensures that $X_s$ and $A_s$ have less features (or edges) so that the explanation highlights important features (or edges) and drop the unimportant ones. One example is to use a differential relaxation of $||A_s||_0 \text{ and } ||X_s||_0$, e.g. a continuous mask~\citep{ying2019gnnexplainer} or $\ell_1$ norm~\citep{luo2020parameterized}, for the size function $s$. On the other hand, $L(X_s, A_s, F)$ penalizes the decay of the model's performance on the subgraph to prevent it from returning trivial solutions. Solving $L(X_s, A_s, F)$ is usually done by gradient descend because this is a non-linear objective w.r.t the node and edge features. Example choices of $L(X_s, A_s, F)$ are mutual information, i.e. GNNExpl~\citep{ying2019gnnexplainer} and PGExpl~\citep{luo2020parameterized}, and Shapley Value, i.e. SubgraphX~\citep{alsentzer2020subgraph}. This paper mainly considers GNNExpl and PGExpl as the evaluation of SubgraphX using the default parameters from DIG~\cite{DIG} exceeds the \texttt{timeout} limit (10 minuteshttps://us06web.zoom.us/j/86767269620 per node) with our current computational resources (Titan RTX) on Cora. We define GNNExpl as follows in Def.~\ref{def: GNNExpl}. 

\begin{definition}[GNNExpl]\label{def: GNNExpl}
Given an input graph $X, A$ and a graph model F, the subgraph explanations are defined as $X_s = X \odot M_1, A_s = A \odot \sigma(M_2)$ where $M_1 \in \{0, 1\}^{n \times d}, M_2 \in \mathbb{R}^{n \times n}$ are solutions to the following objective:
\begin{align}
    \min -\log \text{softmax}[F(X \odot M_1 , A \odot \sigma(M_2))]\\
    s.t. ||M_1||_0 \leq K_1, || \max[\sigma(M_2) - t, 0] ||_0 \leq K_2
\end{align} where $K_1, K_2$ are the maximum number of non-zeros features and edges, respectively, and $0\leq t \leq 1$.
\end{definition}

Since PGExpl shares the same motivation with GNNExpl and only differs on techniques that 1) narrow down the searching space for the optimal $A_s$ and 2) learn a dense layer to jointly explain a set of nodes, we therefore point the readers to \citet{luo2020parameterized} for details. 

\subsection{Attribution Faithfulness}\label{sec:background-attribution-faithfulness}

In this subsection, we firstly describe how faithfulness is defined in the literature for \emph{feature attribution} methods that returns a vector $\Phi$ to explain the feature importance of $x$ given a general function $h(x)$. We discuss how to adapt the faithfulness metric to graph models in Sec.~\ref{sec:general-faithfulness}.

Among properties required for a good explanation, faithfulness ensures the explanation captures the model's behavior with reasonable sensitivity and is used to motivate axioms~\citep{sundararajan2017axiomatic, leino2018influence, sundararajan2020shapley} and evaluations~\citep{NEURIPS2019_a7471fdc, ancona2018towards} for feature attributions. We consider Def.~\ref{def:linear-unfaithfulness} for the faithfulness metric among similar terms in the literature as it provides corresponding characterizations for optimal explanations that minimize the unfaithfulness.

\begin{definition}[Unfaithfulness of Feature Attribution~\citep{NEURIPS2019_a7471fdc}]\label{def:linear-unfaithfulness}
Given a general function $h(x):\mathbb{R}^d \rightarrow \mathbb{R}$ and an attribution vector $\Phi \in \mathbb{R}^d$ for $x$, the unfaithfulness $\Delta$ of $\Phi$ in a neighborhood $B(x, \mathcal{D})$ is $\Delta(w, h, x, \mathcal{D}) = \mathbb{E}_{\epsilon \sim \mathcal{D}} \left[  \Phi^\top \epsilon - (h(x) - h(x - \epsilon)) \right]^2$ and the optimal $\Phi^*$ that minimizes $\phi$ exists as $\Phi^* = \left[\mathbb{E}_{\epsilon \sim \mathcal{D}}(\epsilon \epsilon^\top)\right]^{-1}\cdot \mathbb{E}_{\epsilon \sim \mathcal{D}}(\epsilon(h(x) - h(x - \epsilon))$ if $\mathbb{E}_{\epsilon \sim \mathcal{D}}(\epsilon \epsilon^\top)$ is invertible.
\end{definition}

By by adding a tiny diagonal matrix to $\epsilon \epsilon^\top$, one can ensure that $\mathbb{E}_{\epsilon \sim \mathcal{D}}(\epsilon \epsilon^\top)$ is invertible, which usually leads to trivial affect to the final explanations as detailed in \citet{NEURIPS2019_a7471fdc}. We discuss the examples of distribution $\mathcal{D}$ for graph data in Sec.~\ref{sec:general-faithfulness}.

\section{Faithfulness of Graph Explanations}\label{sec:general-faithfulness}
This section provides a set of characterizations on the faithfulness of graph explanations. Firstly, we generalize Def.~\ref{def:linear-unfaithfulness} to subgraph explanations with \emph{general faithfulness} (Def.~\ref{def:general-unfaithfulness}). Secondly, we show that the limitation of using feature attributions for attributing over edge features. Lastly, we demonstrate that the existing subgraph explanations can lose faithfulness in Prop.~\ref{prop:subgraph-is-not-faithful}. 

% \carlee{This is not entirely clear: do you mean that a straightforward extension of faithfulness from feature attributions to subgraph explanations will not be accurate due to nonlinearity in the subgraph explanations?}

To faithfully capture the local behavior of a deep model $F$, a common approach is to find a surrogate function $p$ to approximate $F$'s output. However, given the ultimate goal of learning $p$ is to understand how $F$ changes in a given neighborhood, one can directly approximate the change of the target function, i.e. $h(x) - h(x-\epsilon)$ in Def.~\ref{def:linear-unfaithfulness}. The following definition therefore characterizes the faithfulness of using an arbitrary function $p$ to explain a graph model $F$ and we discuss how the linearity of $p$ impacts its faithfulness.

\begin{definition}[General Unfaithfulness]\label{def:general-unfaithfulness}
The general unfaithfulness $\Delta(p, F, X, A, \mathcal{D}_X, \mathcal{D}_A)$ (when other variables are clear from the context we write $\Delta(p)$) of using a function $p(X, A)$'s parameters to explain the feature importance of the target graph model $F(X, A)$ in the neighborhood $B(X, \mathcal{D}_X)$ and $B(A, \mathcal{D}_A)$ is defined as follows $\Delta(p) = \mathbb{E}_{\epsilon_X, \epsilon_A} [p(X, A) - p(X-\epsilon_X, A-\epsilon_A) -(F(X, A) - F(X-\epsilon_X, A-\epsilon_A)) ]^2$ where $\epsilon_X \sim \mathcal{D}_X,\epsilon_A \sim \mathcal{D}_A$ and we regard $p$ as a local difference model to $F$.
\end{definition}

Def.~\ref{def:general-unfaithfulness} can be viewed as a measurement on the \emph{difference of difference}; Namely, it is sufficient for $p$ to be a faithful local difference model to $F$ that their first-order differences within a local neighborhood are small. To give some concrete examples of $p$ we will introduce the following cases where we categorize the  distributional influence explanation as using a linear $p$ w.r.t both node and edge features to explain the graph model whereas subgraph explanations are special cases when $p$ is non-linear.

\paragraph{Case I: Linear $p$.} Notice when $p(X, A) = \Phi_X^\top \mathbb{V}(X) + \Phi_A^\top \mathbb{V}(A) $, we have $ p(X, A) - p(X-\epsilon_X, A-\epsilon_A) = \Phi_X^\top \mathbb{V}(\epsilon_X) + \Phi_A^\top \mathbb{V}(\epsilon_A) = [\Phi_X; \Phi_A]^\top [ \mathbb{V}(\epsilon_X);\mathbb{V}(\epsilon_A)]$, Def.~\ref{def:general-unfaithfulness} now reduces to Def.~\ref{def:linear-unfaithfulness}. When $\Phi_X = \triangledown_X F$ and $\Phi_A = \triangledown_A F$, i.e. the Saliency Map of $F$ (Def.~\ref{def:distributional-influence}), $p(X, A)$ can be viewed as a linearization for $F$ at the given input $(X, A)$. Besides, all distributional influences can be viewed as computing linear approximations to $F$ because they are proved to return the weights of $F$ if $F$ is a linear model~\cite{leino2018influence}. We now discuss a limitation of using a linear local difference model $p$. We begin with Prop.~\ref{prop:non-linearity-of-A}.
 
\begin{proposition}\label{prop:non-linearity-of-A}
Given a convolutional graph model $F(X, A)$ with ReLU activations, $F(X, A)$ is locally linear w.r.t $X$ when $A$ is fixed. However, only one of the following statement can hold at one time: 1) $F$ contains one graph convolution layer; 2) when $X$ is fixed, $F$ is not locally linear w.r.t $A$.
\end{proposition}

Prop.~\ref{prop:non-linearity-of-A} suggests a linear $p$ is provably faithful to explain the feature importance of $X$ within an input region where the model is linear, though such regions are usually infinitesimal without robust training strategies, e.g. a GloroNet~\citep{pmlr-v139-leino21a}. On the other hand, Prop.~\ref{prop:non-linearity-of-A} also suggests that a nonlinear local difference function $p$ may be a better fit to attribute over the edge features for graph neural networks with more than two graph convolution layers. In practice, multi-layer graph convolutions are necessary for a variety of tasks because information from 1-hop neighbors are usually not sufficient to predict the node's label. In Sec.~\ref{sec: towards faithfulness} we discuss the way to construct a faithful and nonlinear $p$.

\paragraph{Case II: Nonlinear $p$.} Subgraph explanations (Def.~\ref{def:subgraph-explanation}) are among the use of nonlinear $p$ to model the change of GNN but should be considered as a special case: it adds a special layer at the bottom of the original function $F$, which pre-processes the input, i.e. ablating some features and edges, before feeding them to $F$. Namely, $p(X, A) = F(X_s, A_s)$. Therefore, one necessary condition for $p(X, A)$ to be faithful is that $F$ has exactly the same local behavior within the neighborhood centered at $(X_s, A_s)$ and $(X, A)$. Firstly, subgraph $(X_s, A_s)$ must not change the model's output, which we find is usually not guaranteed. For example, the objective given by Def.~\ref{def: GNNExpl} always maximizes the output probability regardless of the original probability. Secondly, we also hypothesize that when $F$ is less non-linear, e.g. in an early stage of training, subgraph explanations might be more faithful because the decision boundaries in the input space are less complex compared to a well-trained model. Prop.~\ref{prop:subgraph-is-not-faithful} summarizes our analysis, and empirical findings will be discussed in Sec.~\ref{sec:evaluations}.
\begin{proposition}\label{prop:subgraph-is-not-faithful}
Given a local difference model $p(X, A) = F(X_s, A_s)$ defined by a subgraph explanation $X_s, A_s$, and a model $F(X, A)= \{e^\top_vf(X, A)\}_y$ where $f$ is the output of the last layer of graph convolution and $y$ is the target class, the general unfaithfulness $\Delta(p)$ has the following lower-bound $\Delta(p) \geq [C + \mathbb{E}_{\epsilon_X, \epsilon_A} f(X-\epsilon_X, A-\epsilon_A) -
    \mathbb{E}_{\epsilon_X, \epsilon_A}f(X_s-\epsilon_X, A_s-\epsilon_A)]^2$
 where $C = F(X_s, A_s) - F(X, A)$.
\end{proposition}

% Prop.~\ref{prop:subgraph-is-not-faithful} suggests that the constant $C$ is 0 when $F(X_s, A_s)$ outputs the same score as $F(X, A)$. Consider the case for GNNExpl (Def.~\ref{def: GNNExpl}) where the penalty function $L(X_s, A_s)$ minimizes the negative log probability of the prediction for the target class, which does not take the original output $F_j(X, A)$ into account. As a result, minimizing $L(X_s, A_s)$ leads to the maximization of the output probability for the target class, which is not guaranteed to produce a small $C = F(X_s, A_s) - F(X, A)$ unless the original model is very confident for the prediction. Namely, $F(X, A)$ is high enough so that simply increasing $F(X_s, A_s)$ leads to the same gradient update as trying to minimize $C = F(X_s, A_s) - F(X, A)$. On the other hand, when the original model is less certain, i.e. $F_j(X, A)$ is small, Eq.(5) 
% in \tocite{GNN Explainer} may not guarantee to produce faithful subgraph explanations.

% \input{Figure_khop_model}

\section{Towards Faithful Explanations}\label{sec: towards faithfulness}
In this section, we discuss ways to develop a non-linear faithful local difference model $p$ to explain the target GNN. Graph networks can be written as $F(X, A) = A \cdot \Lambda_{M-1} (\cdots (AXW_0) \cdots )W_{M-1}$ where $M$ is the number of graph convolution layers and $\Lambda_*$ is the corresponding activation matrix for ReLUs (we point readers to the proof of Prop.~\ref{prop:non-linearity-of-A} in Appendix.~\ref{appendix:proof} for details in expanding graph networks in matrix products), the non-linearity between $F$ and $A$ is polynomial up to $A^M$ (when all ReLUs are activated). Recall that each entry of the $k$-th power of the adjacency $A^k_{ij}$ indicates the number of paths from $i$ to $j$ with a $k$-hop path, we propose to build $M$ polynomial functions w.r.t $A$ to explain $F(X, A)$. Formally, we introduce Def.~\ref{def:kec}.

\begin{definition}[K-hop Model with a Convolutional Core (KEC)]\label{def:kec}

Suppose $F(X, A)$ has $M$ graph convolution layers, a k-hop local difference model $p(X, A)$ to explain $F$ is defined as $p(X, A) = \sum^M_{k=1} e^\top N(A^k)Xw_k$
% \begin{align}
%     p(X, A) = \sum^M_{k=1} e^\top N(A^k)Xw_k
% \end{align}
where $N(\cdot)$ is the normalization for the adjacency matrix and $w_k (k=1, ..., M)$ are trainable parameters.
\end{definition}
Each component $N(A^k)Xw_k$ in the k-hop model builds a direct edge between a k-hop neighbor and the node of interest. By attributing the output of the $k$-hop local difference mode w.r.t the node features and the edge features, we generate the explanation to $F(X, A)$ as the gradient of $p$, i.e. $\triangledown_X p \text{ and } \triangledown_A p$, for the importance of the nodes and edges, respectively. The analytical expressions of $\triangledown_X p, \triangledown_A p$ are functions of $w_k$ and can be analytically derived using Kronecker product; however, most auto differentiable framework, i.e. Pytorch, can utilize vector-Jacobian tricks to efficiently compute $\triangledown_X p, \triangledown_A p$. The following proposition ensures we can find optimal parameters $w_k$ for the k-hop model to faithfully explain the target GNN.
\begin{proposition}\label{prop:kec}
Suppose $p(X, A)$ is a KEC model for $F(X, A) = {e^\top_vf(X, A)}_y$, the optimal parameters $W = [w^\top_1, w^\top_2, ... w^\top_m]^\top$ that minimizes the general unfaithfulness $\Delta(p)$ around the neighborhood $B(X, \mathcal{D}_X)$ and  $B(A, \mathcal{D}_A)$  is given by $W^* = (\mathbb{E}_{\epsilon_X, \epsilon_A}[\phi \phi^\top])^{-1} \cdot \mathbb{E}_{\epsilon_X, \epsilon_A}[\phi\cdot e^\top_v \delta(f)]$  where $\phi = [\tau^\top_1, \tau_2^\top, ..., \tau_m^\top]^\top, \tau_k =e^\top [N(A^k)X - N((A - \epsilon_A)^k)(X-\epsilon_X)], \delta (f) = f(X, A) - f(X-\epsilon_X, A-\epsilon_A)$ if $\mathbb{E}_{\epsilon_X, \epsilon_A} [\phi\phi^\top]$ is invertible.
\end{proposition}

\paragraph{Complexity} The complexity of solving the optimal parameters using a linear function (Def.~\ref{def:linear-unfaithfulness}) is to invert a symmetric matrix with $n*d+e$ rows where $n$ and $e$ is the number of nodes and edges in the computational graph, and $d$ is the number of node feature dimensions. For KEC, we only need to invert a symmetric matrix with $M*d$ rows where $M$ is the number of graph convolution layers which is usually, e.g. 3 or 4, far less than the number of nodes in practice.
\paragraph{Implementation.} Similar to the linear case (Def.~\ref{def:linear-unfaithfulness}), to inverse $\mathbb{E}_{\epsilon_X, \epsilon_A} [\phi\phi^\top]$ one can add a tiny diagonal matrix or pseudo-inverse instead of actually finding the matrix inverse for approximating the inverse of that expectation. In practice, We have firstly attempted to solve $W^*$ with SVD but run into numerical instability because the smallest eigenvalues of the $\phi$ are too close to 0. We find one way to stabilize the pseudo-inverse is a truncated SVD, which discards the singular vectors if their corresponding singular values are less than a threshold~\cite{SVD}. Empirically we find a threshold of 0.0001 provides a numerical stability for all experiments.

\section{Experiments}\label{sec:evaluations}
This section provides the following empirical studies. Firstly, we measure the general unfaithfulness $\Delta(p)$ of several baseline explanations and the proposed method KEC. We find that KEC is the most faithful method on one synthetic dataset and two real-world datasets when perturbing the node features or edge features in Sec.~\ref{sec:experiment-faithfulness}. Secondly, we demonstrate that gradient-based methods and KEC are both accurate and succinct. Besides, they are also faithful to model's parameters while subgraph explanations, i.e. GNNExpl, may be less sensitive, which serves as an empirical evidence for Prop.~\ref{prop:subgraph-is-not-faithful}. Due to a space limit, the details of this part are included in Appendix.~\ref{appendix:paramter-faithfulness}. Visual comparisons of our approaches are arranged in Fig.~\ref{fig:illustration} and later in Sec.~\ref{sec: case-study} in our case study as this section mainly focuses on quantitative results. 

\subsection{Setup}
\paragraph{Datasets and Models.} We explain the graph model's prediction on the following datasets: 1) BA-Shapes: a synthetic dataset with duplicates of motifs in a house-like shape and random nodes, which has been widely used in the literature. The task is to classify the type of each node being which part of the "house" motif or just random nodes; 2) Cora~\citep{sen:aimag08}: a citation network where nodes represent documents and edges represent citation links and the task is to predict the category of the publication represented by each node; and 3) Citeseer~\citep{sen:aimag08}: a citation network similar to Cora but with more node features. All datasets are trained with graph convolution networks (GCNs). Architecture and parameter details are included in Appendix.~\ref{appendix:gcn-models}.

\paragraph{Baselines.} We include two types of distribution influence in the evaluations: Saliency Map (\textbf{SM}) and Integrated Gradient with a zero vector (\textbf{IG (zero)})  and a random vector (\textbf{IG (random)}) as the baseline and we use 50 points to interpolate points between the baseline and the input, following the frequent choices in the literature. For subgraph explanations, we include: 1) \textbf{GNNExpl} as it is the building block for many follow-up works; 2) a continuous version of GNNExpl (\textbf{GNNExpl (soft)}) where we do not drop edges with small importance scores and we instead consider it as providing a constinous and weighted adjacency matrix $A_{s_{ij}} \leq 1$ and 3) \textbf{PGExpl}. We also include the optimal explanations from Def.~\ref{def:linear-unfaithfulness} (\textbf{Linear}) solved by the truncated SVD. More details on the implementaion of GNNExpl and PGExpl can be found in Appendix~\ref{appendix:gnn-expl}.

\subsection{Explanation Faithfulness}\label{sec:experiment-faithfulness}

\paragraph{Distributions.} The first experiment evaluates the faithfulness of all explanations with Def.~\ref{def:general-unfaithfulness}. Since KEC is motivated to bring non-linearity into the local difference model to faithfully attribute over the edge features, we mainly consider the following distributions: 1) $\mathcal{U}(A, \sigma)$ where the perturbation to each edge weight $A_{ij}$ is uniformly sampled from $[-\sigma, \sigma]$ and we clip each edge weight so that $A_{ij} \geq 0$. We evaluate the case when $\sigma=0.2$ and $0.5$. We consider this case as we drop the edges in a soft way; 2) $\mathcal{B}(A, \sigma)$ where the probability we sample the perturbation from a Bernoulli distribution of the parameter $\sigma$ (consider this case as we drop the edge in a hard way). We evaluate the case when $\sigma=0.2$ and $0.5$.  Our perturbations to edge features include self-loops. Prop.~\ref{prop:kec} does not prevent us from considering perturbations to the node features. Therefore, we also explore  $\mathcal{U}(X, \sigma r)$ where the perturbation to each feature $X_{ij}$ is uniformly sampled from $[-\sigma * r, \sigma * r]$ where $r = \max(X) - \min(X)$ is the range of feature values. In our experiments, $\sigma = 0.2$.

% ; and 2)  mixed distributions that we perturb the node features and edge features simultaneously. We denote this neighborhood as $\mathcal{U}(X, 0.5r) \times \mathcal{U}(A, 0.5)$ for example. This is a relatively larger neighborhood compared to all three distributions above. We do not perturb node features for BA-shapes as they are constants by construction. 
\begin{table*}[t]
   \small
   \centering
   \begin{tabular}{l|ccc|ccc}
   \toprule\toprule
   &\multicolumn{3}{c}{\textbf{BA-shape}} & \multicolumn{3}{c}{\textbf{Cora}} \\ 
   \midrule
   \textbf{Neighborhood} & $\mathcal{U}(X, 0.2r)$ & $\mathcal{U}(A, 0.5)$ & $\mathcal{B}(A, 0.5)$&  $\mathcal{U}(X, 0.2r)$ & $\mathcal{U}(A, 0.5)$ & $\mathcal{B}(A, 0.5)$ \\
   GNNExpl& - & 0.26& 1.29& 0.07& 3.81 & 0.53 \\
   GNNExpl (soft)& - & 0.28&1.35 & 0.02& 0.09 & 0.11  \\
   PGExpl& - &  0.25 & 0.68  &- & 1.35 & 0.75  \\
   SM& - & 0.19 &1.57 &0.02 & 7.26 $\times 10^{-4}$ & 0.38 \\
   IG (zero)& - & 1.59&6.49 &0.03 & 0.21 & 0.74\\
   IG (random)& - & 0.29 &1.48 &0.03 &3.93 $\times 10^{-3}$ \\
   Linear& - & \textbf{0.13}  & \textbf{0.33} & 0.09 &2.3 $\times 10^{-3}$ & 0.45 \\
   KEC& - & \textbf{0.13} &0.50 & \textbf{3.80} $\times \mathbf{10^{-4}}$& \textbf{4.25} $\times \mathbf{10^{-4}}$ & \textbf{0.06}   \\
   \bottomrule
   \end{tabular}
   \caption{General Unfaithfulness $\Delta(p)$ when node and edges are perturbed separately. Results are averaged over 80 nodes for BA-Shapes (we use the same interval [400:700:5] for the node indices selected in the literature~\citep{ying2019gnnexplainer}) and 200 nodes for Cora and Citeseer (see Appendix.~\ref{appendix:citeseer}). Lower scores are better.}
   \label{tab:gf-individual}
\end{table*}

\paragraph{Results.} For BA-shapes, we use node indices \texttt{[400:700:5]} following the choice in the previous work~\citep{ying2019gnnexplainer, luo2020parameterized, PGMExplainer, alsentzer2020subgraph}. For Cora and Citeseer, we use node indices \texttt{[0:1000:5]}. Due to the space limits, results on Citeseer is moved to Appendix~\ref{appendix:citeseer}. To evaluate the general faithfulness $\Delta(p)$ we average 500 points sampled (with the same random seed for all methods) from the distribution of interests. To solve the optimal parameters in Linear and PEC we use 200 samples. 

We average the results on all nodes in Table~\ref{tab:gf-individual}. We make the following observations from the table: 1) KEC is uniformly better than all baselines on real datasets, i.e. Cora and Citeseer, white it matches or is slightly worse than Linear on BA-shapes; 2) Saliency Map (SM) is much faithful than most other baselines for explaining the graph models while Integrated Gradient (IG) is really sensitive to the choice of baselines. A common choice of baseline, a zero vector, from the vision task does not appear to be a reasonable baseline in the graph case. We hypothesize this is because a baseline adjacency matrix with all zeros is too far away from any meaningful given input adjacency matrix. In the image case, moving from a zero baseline to the current image can be viewed as shedding a light on a black image; however, in the graph case, the same intuition does not exist and a random baseline adjacency matrix is much more faithful to explain the local behavior of a graph model than does a zero one. 

Our results highlight that the baseline selection plays an important role if users would like to use IG to explain the graph models; and 4) lastly we find that by interpreting edges with lower importance scores and removing them from the original graph to create a subgraph, i.e. considering GNNExpl instead of GNNExpl (soft), significantly increases the unfaithfulness in the real datasets, i.e. Cora and Citeseer, but not in BA-shapes. This brings our attention to the limitation when using synthetic data to evaluate graph explanations. On the other hand, the difference between GNNExpl and GNNExpl (soft) indicates that edges with lower importance scores are not "noise"; they could be insufficient for the model to make the right decision but are necessary to keep for the purpose of faithful explanations.

In summary, as KEC is motivated to find a non-linear local difference model to faithfully attribute over edges, the current approximation method, truncated SVD, provides reasonable precision to use when evaluating over perturbations on edge features.

\paragraph{Larger Neighborhood, Approximation Error.} We also measure the general unfaithfulness $\Delta p$ for a mixed distribution, i.e. perturbing the node and edge features together. We find that when users want to apply KEC using perturbations on node features, a much larger set of samples are necessary to ensure the faithfulness if the dimensions of node features are significantly high. More experimental details, the relevant table, and our findings are discussed in Appendix.~\ref{appendix:mixed-distribution} and~\ref{appendix:approximation-error}.

\paragraph{Faithfulness to Model Parameters.} So far we develop and evaluate the faithfulness of graph explanations on every single node but being faithful to the target model can also be considered in another context -- that the explanations should be sensitive enough to the model's trainable parameters, i.e. a sanity check~\citep{10.5555/3327546.3327621}. We compare the performance of the explanations against the model's test accuracy with checkpoints saved during training. Experimental details and relevant plots are included in Appendix~\ref{appendix:paramter-faithfulness}) Our results suggest that SoftGNNExpl and PGExpl may not be sensitive enough to reflect the model's performance. That is, these explanations appear to locate relevant edges in the graph even when the model is degenerated. 

\subsection{Efficiency}

Following the efficiency experiments by ~\citet{alsentzer2020subgraph} we also report the timing of our methods and all baselines. Results are shown in Table.~\ref{tab:efficiency} measured on a single Tesla T4. We use the default number of epochs for GNNExpl and PGExpl in the public repository.

\section{Case Study: Anomaly Detection}\label{sec: case-study}
This section conducts a case study of using faithful explanations for real-world GNN applications, combined with domain-specific knowledge, to help interpret explanation results.
We use the dataset collected from a real-world secure water treatment system (SWAT) \citep{goh2016swat-dataset} to build a GNN model for anomaly detection in multivariate time series data. Prior work~\citep{deng2021swat-aaai} has demonstrated GNNs' efficacy in such a task, however, we want to provide more insights into how GNN models make their prediction with our explanation methods. Since many methods (e.g., GNNExpl) are originally designed only for classification tasks, we will use visual comparisons in lieu of formal quantitative analysis in this case study. The SWAT dataset~\citep{goh2016swat-dataset} contains time series readings of 51 nodes (sensors and actuators) in a water treatment testbed over 11 days. To generate anomaly, researchers manually temper with individual nodes (e.g., shutdown) or spoof readings and label the corresponding data points as being under attack. The training procedure for anomaly detection with a GNN together with the quantity of interest we use to attribute over the node and edge features are included Appendix~\ref{appendix:swat} . We compare KEC and Saliency Map, the faithful ones, with GNNExpl (soft), the less faithful one, found in Sec.~\ref{sec:evaluations}.

% Original Zifan's text below, for reference
% This section does not serve as an evaluation but as a case study of using faithful explanations for real-world applications with GNNs, where more domain knowledge is available to help interpret the explanation results. Since PGExpl and GNNExpl\footnote{An unofficial extension for GNNExpl to the regression task with a different target function is available on Pytorch Geometric. We will use this implementation as a baseline in visual comparison instead of a formal baseline in more quantitative evaluation.} are not original motivated for tasks beyond classifications, we do not perform quantitative evaluations. \outline{todo: A short justification of why anomaly detection is an interesting case.} We choose the SWAT water treatment dataset~\cite{goh2016swat-dataset} where time-series readings of 51 node (sensors and actuators) in a water treatment testbed are collected and used as node features. The target task for SWAT is a node-level anomaly detection, i.e. the malfunction of sensors caused by malicious shutdown or spoofed readings. Recent work has shown GNN is able to provide the state-of-the-art performance on reporting anomalous behaviors in SWAT sensors~\tocite{}. In this section, we firstly build a graph model to detect anomalous nodes in SWAT and visualize explanations to understand the model's prediction (implementation details to follow in Appendix~\ref{appendix:swat}). 
% Secondly, we visualize KEC, SM, the most faithful methods, and a less faithful one, GNNExpl, found in Sec.~\ref{sec:evaluations}. 

\begin{figure}[t]
\centering
    \begin{subfigure}[b]{0.5\textwidth}
        \centering
        \includegraphics[width=\columnwidth]{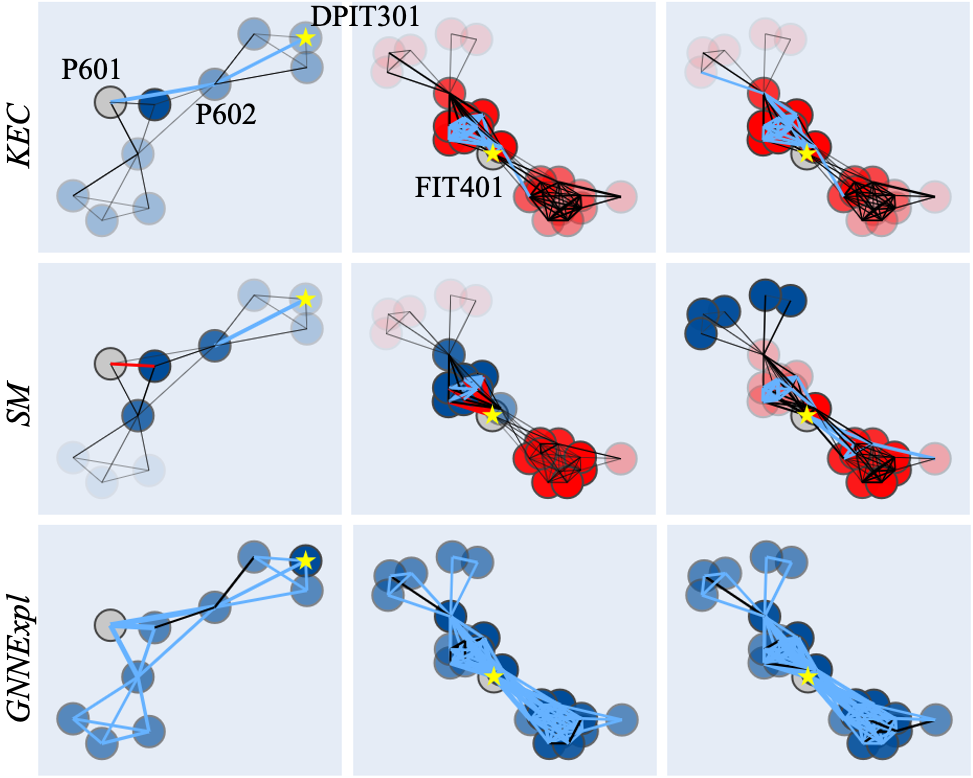}
        \subcaption{}\label{fig:swat-visualization}
    \end{subfigure}%
    \hfill
    \begin{subfigure}[b]{0.45\textwidth}
      \small
      \centering
      \resizebox{\columnwidth}{!}{\begin{tabular}{l|ccc}
      \toprule\toprule
      & \text{BA-shapes} & \text{Cora} & \text{Citeseer} \\ 
      \# of GCN layers& 3 & 2& 2 \\ 
      \midrule
      GNNExpl (100 epochs) & 0.70 & 0.50	& 0.53\\
      PGExpl (30 epochs) & 0.62&	0.45	&0.47\\
      SM & 0.01	&0.01	&0.07\\
      IG (50 samples) & 0.02	&0.45	&1.45\\
      Linear (200 samples) & 1.01 $\times 10^{-3}$ 	&0.01	&0.04\\
      Linear (1500 samples)& 1.17 $\times 10^{-3}$	&0.01	&0.04\\
      KEC (200 samples) & 0.04	&0.03	&0.06\\
      KEC (1500 samples) & 0.04	&0.03	&0.06\\
      \bottomrule
      \end{tabular}}
      \subcaption{} \label{tab:efficiency}
    \end{subfigure}%
    \caption{(a) Visualization of graph explanations for SWAT anomaly detection using different methods for 3 attacks (left to right). Grey nodes are abnormal nodes reported by GNN (P601, FIT401, FIT401), while yellow stars are nodes under attack in ground truth (DPIT301, FIT401, FIT401). The importance of nodes and edges is colored with red (positive) and blue (negative) and their magnitudes are shown by opacity (for nodes) and width (for edges). (b) Per node running time (seconds) measured on a Tesla T4. Training times given for GNNExpl and PGExpl. Results are averaged over 100 nodes. }
\end{figure}

% \begin{figure}[t]

% % \subcaptionbox{KEC}{        
% %     \includegraphics[width=0.33\columnwidth]{swat_images/pec1.png}
% %     \includegraphics[width=0.33\columnwidth]{swat_images/pec2.png}
% %     \includegraphics[width=0.33\columnwidth]{swat_images/pec3.png}
% % }

% % \subcaptionbox{Saliency Map}{
% %     \includegraphics[width=0.33\columnwidth]{swat_images/sm1.png}
% %     \includegraphics[width=0.33\columnwidth]{swat_images/sm2.png}
% %     \includegraphics[width=0.33\columnwidth]{swat_images/sm3.png}
% % }

% % \subcaptionbox{GNN Explainer}{
% %     \includegraphics[width=0.33\columnwidth]{swat_images/gnn1.png}
% %     \includegraphics[width=0.33\columnwidth]{swat_images/gnn2.png}
% %     \includegraphics[width=0.33\columnwidth]{swat_images/gnn3.png}
% % }
% \centering
% \includegraphics[width=0.5\columnwidth]{swat_images/SWAT.png}
   
% \caption{Visualization of graph explanations for SWAT anomaly detection using different methods for 3 attacks (left to right). Grey nodes are abnormal nodes reported by GNN (P601, FIT401, FIT401), while yellow stars are nodes under attack in ground truth (DPIT301, FIT401, FIT401). The importance of nodes and edges is colored with red (positive) and blue (negative) and their magnitudes are shown by opacity (for nodes) and width (for edges). }
% \label{fig:swat-visualization}
% \end{figure}

\paragraph{Findings.} 
\Cref{fig:swat-visualization} shows the visualization of three attack instances explained by different methods. Images of the same column are from the same attack. The actual node under attack is marked by yellow starts while the model's predictions are colored in gray. Blue nodes and edges are positive influences --- what makes the model believe the gray node has an anomalous reading --- while red ones mean negative. Among the three attacks, we find several interesting insights: (1) The model incorrectly attributes nodes for attack \#1. The attack spoof the pressure sensor (DPIT301) readings to repeatedly trigger the backwash process (controlled by pump P602). As a result, the model predicts a correlated pump (P601) should also have a low speed reading well below its normal operating range. Looking at the visualizations, KEC clearly shows the influence of DPIT301--P602--P601 with blue edges. (2) Both attack \#2 and \#3 directly set flow meter FIT401's value as 0 to shut down subsequent processes. Since they are the same attacks, their explanations should reflect the similarity as well. Surprisingly, SM generates inconsistent explanations for these two attacks. We find the disagreement mainly focuses on the top left group w.r.t the stared node, which provides us a direction to investigate the model's behavior. (3) Overall, GNNExpl's results might be less insightful since almost everything is important to the prediction and a lack of anti-significant edges and nodes. We believe this is because GNNExpl uses a sigmoid function at the top of the mask (Def.~\ref{def: GNNExpl}), which can i) easily push the significance scores towards either 0 and 1; and ii) flip the sign of negative scores.

\section{Related Work}~\label{sec:related-work}
\paragraph{Faithfulness in General.} We first compare terms used in the literature that are similar to \emph{general faithfulness}. \citet{NEURIPS2019_a7471fdc} use INFD score to quantify the faithfulness for gradient-based attributions, which is one of the building block for our metric. Another line of work is the ablation study for input features~\citep{ancona2018towards, Wang2020InterpretingIO}, which can be viewed as evaluating whether the explanation is faithful when compared with a baseline vector, e.g. all zeros, where \emph{general faithfulness} can be used for any neighborhood of interest as long as the explanation can be written as a local difference model $p$. To ensure the explanation captures the local behavior of the target function, LIME~\cite{lime} instead minimizes the difference between a surrogate model and the target function for neighbors, where the difference is also weighted by the distance. We highlight that our methods are different from LIME in the following aspects: 1) LIME considers that the surrogate model takes points in a space different from the GNN's input space for the purpose of interpretability; 2) LIME requires that the surrogate model's output becomes similar to the target function, whereas \emph{general faithfulness} requires the change of $p$ to align with the change of the target GNN, i.e. the difference of difference. 3) LIME uses a linear model and enforce sparsity of the weights, where KEC uses k-hop models and considers non-linearity for the adjacency matrices. We argue that the sparsity penalty can become dominant compared to faithfulness, resulting in non-faithful but sparse explanations. The graph variation of LIME, GraphLIME~\citep{huang2020graphlime}, uses Hilbert-Schmidt Independence Criterion Lasso to encoder non-linear interactions between node features. GraphLIME can not attribute over edges and therefore are not compared in this paper.

\paragraph{Faithfulness in Graph Models}. The most similar metric proposed for the faithfulness of grpah explanation is \emph{fidelity} by \cite{alsentzer2020subgraph}, which measures the difference of the model's output when the input is replaced with the importance subgraph. \emph{fidelity} therefor can be considered as a special case for \emph{general faithfulness} when the neighborhood of interest is only the input itself. \citet{NEURIPS2020_417fbbf2} introduces four metrics for attributions among which the term \emph{faithfulness} refers to how explanations change when the training set is perturbed. i.e. introducing noises and permuting the labels. \citet{NEURIPS2020_417fbbf2} finds IG and GradCAM~\cite{Gradcam} is more correlated with the model's performance, which agrees with our conclusions in Sec.~\ref{sec:evaluations} from another way of varying the model's performance. Using checkpoints may provide more continuous measurement than training multiple models because we are continuously monitoring the same network. Besides, our experiments also cover subgraph explanations that are not included by \citet{NEURIPS2020_417fbbf2}.

\paragraph{Explaining Graph Classifications.} Another line of related in work is to explain graph classifications, where all node and edge features are used and the task is to attribute the model's output over the entire graph. One simple way to adapt KEC explanation into this case is by modifying the $e_v$ vector (Sec.~\ref{sec:background-attribution-faithfulness}) to be a vector of all ones, which should function as a global sum pooling operation and Prop.~\ref{prop:kec} still holds in this case. As the paper mainly focuses on explaining node classification and its related tasks, e.g. node-level anomaly detection, which already covers a large scope of research topics, we leave this as a future work to follow up with.
\section{Conclusion}\label{sec:conclusion}

In this paper, we study the problem, \emph{how should one construct and evaluate faithful explanations for GNNs?}. We build a general faithfulness metric, \emph{general faithfulness}, to both subgraph explanations and feature attributions methods to understand how well these approaches capture the model's behavior in a local neighborhood. In particular, we introduce KEC as a faithful explanation to attribute a GNN's output w.r.t node and edge features. Our empirical studies over a set of existing explanations methods highlight cases when they may not be faithful. Finally, we provide a case study on anomaly detection for the use case of faithful explanations in debugging and improving the model's behavior beyond classification tasks with GNNs.

%%%%%%%%%%%%%%%%%%%%%%%%%%%%%%%%%%%%%%%%%%%%%%%%%%%%%%%%%%%%

\bibliography{ref}

\begin{thebibliography}{37}
\providecommand{\natexlab}[1]{#1}
\providecommand{\url}[1]{\texttt{#1}}
\expandafter\ifx\csname urlstyle\endcsname\relax
  \providecommand{\doi}[1]{doi: #1}\else
  \providecommand{\doi}{doi: \begingroup \urlstyle{rm}\Url}\fi

\bibitem[Adebayo et~al.(2018)Adebayo, Gilmer, Muelly, Goodfellow, Hardt, and
  Kim]{10.5555/3327546.3327621}
Julius Adebayo, Justin Gilmer, Michael Muelly, Ian Goodfellow, Moritz Hardt,
  and Been Kim.
\newblock Sanity checks for saliency maps.
\newblock In \emph{Proceedings of the 32nd International Conference on Neural
  Information Processing Systems}, page 9525–9536, Red Hook, NY, USA, 2018.
  Curran Associates Inc.

\bibitem[Alsentzer et~al.(2020)Alsentzer, Finlayson, Li, and
  Zitnik]{alsentzer2020subgraph}
Emily Alsentzer, Samuel~G Finlayson, Michelle~M Li, and Marinka Zitnik.
\newblock Subgraph neural networks.
\newblock In \emph{Proceedings of Neural Information Processing Systems,
  NeurIPS}, 2020.

\bibitem[Ancona et~al.(2018)Ancona, Ceolini, Öztireli, and
  Gross]{ancona2018towards}
Marco Ancona, Enea Ceolini, Cengiz Öztireli, and Markus Gross.
\newblock Towards better understanding of gradient-based attribution methods
  for deep neural networks.
\newblock In \emph{International Conference on Learning Representations}, 2018.
\newblock URL \url{https://openreview.net/forum?id=Sy21R9JAW}.

\bibitem[Chen et~al.(2022)Chen, Zhang, You, Zheng, and Lambotharan]{9462385}
Tianrui Chen, Xinruo Zhang, Minglei You, Gan Zheng, and Sangarapillai
  Lambotharan.
\newblock A gnn-based supervised learning framework for resource allocation in
  wireless iot networks.
\newblock \emph{IEEE Internet of Things Journal}, 9\penalty0 (3):\penalty0
  1712--1724, 2022.
\newblock \doi{10.1109/JIOT.2021.3091551}.

\bibitem[Deng and Hooi(2021)]{deng2021swat-aaai}
Ailin Deng and Bryan Hooi.
\newblock Graph neural network-based anomaly detection in multivariate time
  series.
\newblock In \emph{Proceedings of the AAAI Conference on Artificial
  Intelligence}, volume~35, pages 4027--4035, 2021.

\bibitem[Fan et~al.(2019)Fan, Ma, Li, He, Zhao, Tang, and Yin]{fan2019graph}
Wenqi Fan, Yao Ma, Qing Li, Yuan He, Eric Zhao, Jiliang Tang, and Dawei Yin.
\newblock Graph neural networks for social recommendation.
\newblock In \emph{The World Wide Web Conference}, pages 417--426. ACM, 2019.

\bibitem[Fey and Lenssen(2019)]{Fey/Lenssen/2019}
Matthias Fey and Jan~E. Lenssen.
\newblock Fast graph representation learning with {PyTorch Geometric}.
\newblock In \emph{ICLR Workshop on Representation Learning on Graphs and
  Manifolds}, 2019.

\bibitem[Fong and Vedaldi(2017)]{Fong2017InterpretableEO}
Ruth~C. Fong and Andrea Vedaldi.
\newblock Interpretable explanations of black boxes by meaningful perturbation.
\newblock \emph{2017 IEEE International Conference on Computer Vision (ICCV)},
  pages 3449--3457, 2017.

\bibitem[Fromherz et~al.(2021)Fromherz, Leino, Fredrikson, Parno, and
  Pasareanu]{fromherz2021fast}
Aymeric Fromherz, Klas Leino, Matt Fredrikson, Bryan Parno, and Corina
  Pasareanu.
\newblock Fast geometric projections for local robustness certification.
\newblock In \emph{International Conference on Learning Representations}, 2021.
\newblock URL \url{https://openreview.net/forum?id=zWy1uxjDdZJ}.

\bibitem[Goh et~al.(2016)Goh, Adepu, Junejo, and Mathur]{goh2016swat-dataset}
Jonathan Goh, Sridhar Adepu, Khurum~Nazir Junejo, and Aditya Mathur.
\newblock A dataset to support research in the design of secure water treatment
  systems.
\newblock In \emph{International conference on critical information
  infrastructures security}, pages 88--99. Springer, 2016.

\bibitem[Hooker et~al.(2019)Hooker, Erhan, Kindermans, and Kim]{Hooker2019ABF}
Sara Hooker, D.~Erhan, Pieter-Jan Kindermans, and Been Kim.
\newblock A benchmark for interpretability methods in deep neural networks.
\newblock In \emph{NeurIPS}, 2019.

\bibitem[Huang et~al.(2021)Huang, Xu, Xu, Dai, Xia, Lu, Bo, Xing, Lai, and
  Ye]{Huang2021KnowledgeawareCG}
Chao Huang, Huance Xu, Yong Xu, Peng Dai, Lianghao Xia, Mengyin Lu, Liefeng Bo,
  Hao Xing, Xiaoping Lai, and Yanfang Ye.
\newblock Knowledge-aware coupled graph neural network for social
  recommendation.
\newblock \emph{ArXiv}, abs/2110.03987, 2021.

\bibitem[Huang et~al.(2020)Huang, Yamada, Tian, Singh, Yin, and
  Chang]{huang2020graphlime}
Qiang Huang, Makoto Yamada, Yuan Tian, Dinesh Singh, Dawei Yin, and Yi~Chang.
\newblock Graphlime: Local interpretable model explanations for graph neural
  networks, 2020.

\bibitem[Jordan et~al.(2019)Jordan, Lewis, and
  Dimakis]{10.5555/3454287.3455548}
Matt Jordan, Justin Lewis, and Alexandros~G. Dimakis.
\newblock \emph{Provable Certificates for Adversarial Examples: Fitting a Ball
  in the Union of Polytopes}.
\newblock Curran Associates Inc., Red Hook, NY, USA, 2019.

\bibitem[Kipf and Welling(2017)]{kipf2017semi}
Thomas~N. Kipf and Max Welling.
\newblock Semi-supervised classification with graph convolutional networks.
\newblock In \emph{International Conference on Learning Representations
  (ICLR)}, 2017.

\bibitem[Leino et~al.(2018)Leino, Sen, Datta, Fredrikson, and
  Li]{leino2018influence}
Klas Leino, Shayak Sen, Anupam Datta, Matt Fredrikson, and Linyi Li.
\newblock Influence-directed explanations for deep convolutional networks.
\newblock In \emph{2018 IEEE International Test Conference (ITC)}, pages 1--8.
  IEEE, 2018.

\bibitem[Leino et~al.(2021)Leino, Wang, and Fredrikson]{pmlr-v139-leino21a}
Klas Leino, Zifan Wang, and Matt Fredrikson.
\newblock Globally-robust neural networks.
\newblock In Marina Meila and Tong Zhang, editors, \emph{Proceedings of the
  38th International Conference on Machine Learning}, volume 139 of
  \emph{Proceedings of Machine Learning Research}, pages 6212--6222. PMLR,
  18--24 Jul 2021.

\bibitem[Liu et~al.(2021)Liu, Luo, Wang, Xie, Yuan, Gui, Yu, Xu, Zhang, Liu,
  Yan, Liu, Fu, Oztekin, Zhang, and Ji]{DIG}
Meng Liu, Youzhi Luo, Limei Wang, Yaochen Xie, Hao Yuan, Shurui Gui, Haiyang
  Yu, Zhao Xu, Jingtun Zhang, Yi~Liu, Keqiang Yan, Haoran Liu, Cong Fu, Bora~M
  Oztekin, Xuan Zhang, and Shuiwang Ji.
\newblock {DIG}: A turnkey library for diving into graph deep learning
  research.
\newblock \emph{Journal of Machine Learning Research}, 22\penalty0
  (240):\penalty0 1--9, 2021.
\newblock URL \url{http://jmlr.org/papers/v22/21-0343.html}.

\bibitem[Lundberg and Lee(2017)]{NIPS2017_7062}
Scott~M Lundberg and Su-In Lee.
\newblock A unified approach to interpreting model predictions.
\newblock In I.~Guyon, U.~V. Luxburg, S.~Bengio, H.~Wallach, R.~Fergus,
  S.~Vishwanathan, and R.~Garnett, editors, \emph{Advances in Neural
  Information Processing Systems 30}, pages 4765--4774. Curran Associates,
  Inc., 2017.
\newblock URL
  \url{http://papers.nips.cc/paper/7062-a-unified-approach-to-interpreting-model-predictions.pdf}.

\bibitem[Luo et~al.(2020)Luo, Cheng, Xu, Yu, Zong, Chen, and
  Zhang]{luo2020parameterized}
Dongsheng Luo, Wei Cheng, Dongkuan Xu, Wenchao Yu, Bo~Zong, Haifeng Chen, and
  Xiang Zhang.
\newblock Parameterized explainer for graph neural network.
\newblock \emph{Advances in Neural Information Processing Systems}, 33, 2020.

\bibitem[Mu et~al.(2019)Mu, Zha, He, and Tang]{Mu2019GraphAN}
Nan Mu, Daren Zha, Yuanye He, and Zhihao Tang.
\newblock Graph attention networks for neural social recommendation.
\newblock \emph{2019 IEEE 31st International Conference on Tools with
  Artificial Intelligence (ICTAI)}, pages 1320--1327, 2019.

\bibitem[Petsiuk et~al.(2018)Petsiuk, Das, and Saenko]{Petsiuk2018RISERI}
Vitali Petsiuk, Abir Das, and Kate Saenko.
\newblock Rise: Randomized input sampling for explanation of black-box models.
\newblock In \emph{BMVC}, 2018.

\bibitem[Ribeiro et~al.(2016)Ribeiro, Singh, and Guestrin]{lime}
Marco~Tulio Ribeiro, Sameer Singh, and Carlos Guestrin.
\newblock "why should {I} trust you?": Explaining the predictions of any
  classifier.
\newblock In \emph{Proceedings of the 22nd {ACM} {SIGKDD} International
  Conference on Knowledge Discovery and Data Mining, San Francisco, CA, USA,
  August 13-17, 2016}, pages 1135--1144, 2016.

\bibitem[Romberg(1999)]{SVD}
J.~Romberg.
\newblock Stable reconstruction with the truncated svd, 1999.
\newblock URL
  \url{https://cpn-us-w2.wpmucdn.com/sites.gatech.edu/dist/2/436/files/2017/07/17-notes-6250-f16.pdf}.

\bibitem[Sanchez-Lengeling et~al.(2020)Sanchez-Lengeling, Wei, Lee, Reif, Wang,
  Qian, McCloskey, Colwell, and Wiltschko]{NEURIPS2020_417fbbf2}
Benjamin Sanchez-Lengeling, Jennifer Wei, Brian Lee, Emily Reif, Peter Wang,
  Wesley Qian, Kevin McCloskey, Lucy Colwell, and Alexander Wiltschko.
\newblock Evaluating attribution for graph neural networks.
\newblock In \emph{Advances in Neural Information Processing Systems},
  volume~33, 2020.

\bibitem[Selvaraju et~al.(2019)Selvaraju, Cogswell, Das, Vedantam, Parikh, and
  Batra]{Gradcam}
Ramprasaath~R. Selvaraju, Michael Cogswell, Abhishek Das, Ramakrishna Vedantam,
  Devi Parikh, and Dhruv Batra.
\newblock Grad-cam: Visual explanations from deep networks via gradient-based
  localization.
\newblock \emph{International Journal of Computer Vision}, 128\penalty0
  (2):\penalty0 336–359, Oct 2019.
\newblock ISSN 1573-1405.
\newblock \doi{10.1007/s11263-019-01228-7}.
\newblock URL \url{http://dx.doi.org/10.1007/s11263-019-01228-7}.

\bibitem[Sen et~al.(2008)Sen, Namata, Bilgic, Getoor, Gallagher, and
  Eliassi-Rad]{sen:aimag08}
Prithviraj Sen, Galileo~Mark Namata, Mustafa Bilgic, Lise Getoor, Brian
  Gallagher, and Tina Eliassi-Rad.
\newblock Collective classification in network data.
\newblock \emph{AI Magazine}, 29\penalty0 (3):\penalty0 93--106, 2008.

\bibitem[Simonyan et~al.(2013)Simonyan, Vedaldi, and
  Zisserman]{simonyan2013deep}
Karen Simonyan, Andrea Vedaldi, and Andrew Zisserman.
\newblock Deep inside convolutional networks: Visualising image classification
  models and saliency maps, 2013.

\bibitem[Smilkov et~al.(2017)Smilkov, Thorat, Kim, Viégas, and
  Wattenberg]{smilkov2017smoothgrad}
Daniel Smilkov, Nikhil Thorat, Been Kim, Fernanda Viégas, and Martin
  Wattenberg.
\newblock Smoothgrad: removing noise by adding noise, 2017.

\bibitem[Sundararajan and Najmi(2020)]{sundararajan2020shapley}
Mukund Sundararajan and Amir Najmi.
\newblock The many shapley values for model explanation, 2020.

\bibitem[Sundararajan et~al.(2017)Sundararajan, Taly, and
  Yan]{sundararajan2017axiomatic}
Mukund Sundararajan, Ankur Taly, and Qiqi Yan.
\newblock Axiomatic attribution for deep networks.
\newblock In \emph{Proceedings of the 34th International Conference on Machine
  Learning-Volume 70}, pages 3319--3328. JMLR. org, 2017.

\bibitem[Vu and Thai(2020)]{PGMExplainer}
Minh~N. Vu and My~T. Thai.
\newblock Pgm-explainer: Probabilistic graphical model explanations for graph
  neural networks.
\newblock In \emph{Advances in Neural Information Processing Systems}, 2020.

\bibitem[Wang et~al.(2020{\natexlab{a}})Wang, Wang, Du, Yang, Zhang, Ding,
  Mardziel, and Hu]{wang2020score}
Haofan Wang, Zifan Wang, Mengnan Du, Fan Yang, Zijian Zhang, Sirui Ding, Piotr
  Mardziel, and Xia Hu.
\newblock Score-cam: Score-weighted visual explanations for convolutional
  neural networks.
\newblock In \emph{Proceedings of the IEEE/CVF conference on computer vision
  and pattern recognition workshops}, pages 24--25, 2020{\natexlab{a}}.

\bibitem[Wang et~al.(2020{\natexlab{b}})Wang, Mardziel, Datta, and
  Fredrikson]{Wang2020InterpretingIO}
Zifan Wang, Piotr Mardziel, Anupam Datta, and Matt Fredrikson.
\newblock Interpreting interpretations: Organizing attribution methods by
  criteria.
\newblock \emph{2020 IEEE/CVF Conference on Computer Vision and Pattern
  Recognition Workshops (CVPRW)}, pages 48--55, 2020{\natexlab{b}}.

\bibitem[Yao and Joe-Wong(2022)]{yao2022fedgcn}
Yuhang Yao and Carlee Joe-Wong.
\newblock Fedgcn: Convergence and communication tradeoffs in federated training
  of graph convolutional networks, 2022.

\bibitem[Yeh et~al.(2019)Yeh, Hsieh, Suggala, Inouye, and
  Ravikumar]{NEURIPS2019_a7471fdc}
Chih-Kuan Yeh, Cheng-Yu Hsieh, Arun Suggala, David~I Inouye, and Pradeep~K
  Ravikumar.
\newblock On the (in)fidelity and sensitivity of explanations.
\newblock In \emph{Advances in Neural Information Processing Systems}, 2019.

\bibitem[Ying et~al.(2019)Ying, Bourgeois, You, Zitnik, and
  Leskovec]{ying2019gnnexplainer}
Rex Ying, Dylan Bourgeois, Jiaxuan You, Marinka Zitnik, and Jure Leskovec.
\newblock Gnnexplainer: Generating explanations for graph neural networks.
\newblock \emph{Advances in neural information processing systems},
  32:\penalty0 9240, 2019.

\end{thebibliography}
\bibliographystyle{plainnat}

\newpage
\appendix

\newpage
\section{Proofs}
\label{appendix:proof}

\textbf{Proposition}~\ref{prop:non-linearity-of-A}
Given a convolutional graph model $F(X, A)$ with ReLU activations, $F(X, A)$ is locally linear w.r.t $X$ when $A$ is fixed. However, only one of the following statement can hold at one time: 1) $F$ contains only one layer of graph convolution; 2) when $X$ is fixed, $F$ is not a local linear function w.r.t $A$.

\begin{proof}

It is easy to find when $F$ has only one layer of graph convolution, it can be written as $F(X, A) = ReLU(N(A)XW)$ where $N(A)$ is the normalized adjacency matrix and we omit the bias. Because ReLU is piece-wise linear, $F$ is also piece-wise linear w.r.t $X$ and $A$. We now consider a two-layer graph convolution network
 \begin{align}
     F(X, A) = ReLU(N(A)\cdot ReLU(N(A)XW_0)\cdot W_1)
 \end{align}
 Consider a set of points $P = \{(X_i, A_i)\}$ in the input space, if the activation status, \texttt{OFF} or \texttt{ON}, of all ReLUs remain the same for these points, we refer $P$ as an \emph{activation region} in the input space and write
 \begin{align}
     \forall (X_i, A_i) \in P, F(X, A) = \Lambda_1(N(A)\cdot\Lambda_0(N(A)XW_0)\cdot W_1)
 \end{align}
where $\Lambda_*$ are tensors with the diagonals of the last two dimensions being 0 or 1, indicating whether the pre-activations of ReLUs are positive or not. The convexity of $P$ follows from observations~\cite{10.5555/3454287.3455548, fromherz2021fast}. We now make the following observations. Suppose $A$ is fixed, $\forall (X_i, A) \in P$, $F$ is a linear function w.r.t $X$. In this case, a linear $p=\mathbb{V}((\triangledown_X F))^\top \mathbb{V}(X)$ is always able to faithfully capture the change of $F$ within the activation region $P$. However, when $X$ is fixed, F is still a non-linear function w.r.t $A$ within most activation regions \footnote{Except some special cases, e.g. $\Lambda_0, \Lambda_1$ being zeros tensors.}. It is easy to find that $F$ can be locally linear w.r.t. $A$ contains only if and only if the number of graph convolutions is 1. The analysis for a network with more than two convolutional layers is similar.

\end{proof}

\textbf{Proposition~\ref{prop:subgraph-is-not-faithful}}
Given a local difference model $p(X, A) = F(X_s, A_s)$ defined by a subgraph explanation $X_s, A_s$, and a model $F(X, A)= \{e^\top_vf(X, A)\}_y$ where $f$ is the output of the last layer of graph convolution and $y$ is the target class, the general unfaithfulness $\Delta(p)$ has the following lower-bound $\Delta(p) \geq [C + \mathbb{E}_{\epsilon_X, \epsilon_A} f(X-\epsilon_X, A-\epsilon_A) -
    \mathbb{E}_{\epsilon_X, \epsilon_A}f(X_s-\epsilon_X, A_s-\epsilon_A)]^2$
 where $C = F(X_s, A_s) - F(X, A)$.

\begin{proof}
By the definition of the general unfaithfulness, we have 
\begin{align}
    \Delta(p) = \mathbb{E}_{\epsilon_X, \epsilon_A} [f(X_s, A_s) - f(X_s-\epsilon_X, A_s-\epsilon_A) \\
    - (f(X, A) - f(X-\epsilon_X, A-\epsilon_A))]^2
\end{align} Because $(\cdot)^2$ is a convex function, we exchange the order of expectation with the square function and finds a lowe-bound of $\Delta(p)$ by Jensen's Inequality as 
\begin{align}
\Delta(p) &\geq [\mathbb{E}_{\epsilon_X, \epsilon_A} [f(X_s, A_s) - f(X_s-\epsilon_X, A_s-\epsilon_A) - \\ &(f(X, A) - f(X-\epsilon_X, A-\epsilon_A))]]^2\\
&\geq [\mathbb{E}_{\epsilon_X, \epsilon_A} [f(X_s, A_s) - f(X, A) + \\ & f(X-\epsilon_X, A-\epsilon_A) - f(X_s-\epsilon_X, A_s-\epsilon_A)]]^2\\
&\geq [C + \mathbb{E}_{\epsilon_X, \epsilon_A} f(X-\epsilon_X, A-\epsilon_A) \\ &-
    \mathbb{E}_{\epsilon_X, \epsilon_A}f(X_s-\epsilon_X, A_s-\epsilon_A)]^2
\end{align} where  $C = F(X_s, A_s) - F(X, A)$.
\end{proof}

\textbf{Proposition~\ref{prop:kec}}
Suppose $p(X, A)$ is a KEC model for $F(X, A) = {e^\top_vf(X, A)}_y$, the optimal parameters $W = [w^\top_1, w^\top_2, ... w^\top_m]^\top$ that minimizes the general unfaithfulness $\Delta(p)$ around the neighborhood $B(X, \mathcal{D}_X)$ and  $B(A, \mathcal{D}_A)$  is given by $W^* = (\mathbb{E}_{\epsilon_X, \epsilon_A}[\phi \phi^\top])^{-1} \cdot \mathbb{E}_{\epsilon_X, \epsilon_A}[\phi\cdot e^\top_v \delta(f)]$  where $\phi = [\tau^\top_1, \tau_2^\top, ..., \tau_m^\top]^\top, \tau_k =e^\top [N(A^k)X - N((A - \epsilon_A)^k)(X-\epsilon_X)], \delta (f) = f(X, A) - f(X-\epsilon_X, A-\epsilon_A)$ if $\mathbb{E}_{\epsilon_X, \epsilon_A} [\phi\phi^\top]$ is invertible.

\begin{proof}

By Def.~\ref{def:general-unfaithfulness}, we write $\Delta(p)$ for $p(X, A) = \sum^M_{k=1} e^\top_vN(A^k)Xw_k$ as follows: $\Delta(p) = \mathbb{E}_{\epsilon_X, \epsilon_A} [\sum^M_{k=1} e^\top N(A^k)Xw_k - (\sum^M_{k=1} e^\top N((A-\epsilon_A)^k)((X-\epsilon_X)w_k) -(F(X, A) - F(X-\epsilon_X, A-\epsilon_A)) ]^2$ 
Firstly, we deal with $$\sum^M_{k=1} e^\top N(A^k)Xw_k - (\sum^M_{k=1} e^\top N((A-\epsilon_A)^k)((X-\epsilon_X)w_k)$$
which can be simplified as $\sum^M_{k=1} \tau_k \cdot w_k$ where $\tau_k = e^\top[N(A^k)X - N((A-\epsilon_A)^k)(X-\epsilon_X)]$. By introducing the following matrices:
\begin{align}
    W = [w^\top_1, w^\top_2, ... w^\top_m]^\top \in \mathbb{R}^{Md}, \phi = [\tau, \tau_2, ..., \tau_m] \in \mathbb{R}^{Md}
\end{align} where $M$ is the number of convolutional layers and $d$ is the number of node features, we re-denote 
\begin{align}
     \phi^\top W = \sum^M_{k=1} N(A^k)Xw_k - (\sum^M_{k=1} N((A-\epsilon_A)^k)((X-\epsilon_X)w_k)
\end{align}

With the definition of $F(X, A) = {e^\top_vf(X, A)}_y$, we re-denote
\begin{align}
    F(X, A) - F(X-\epsilon_X, A-\epsilon_A) = e^\top_v \delta(f)
\end{align} where $\delta(f) = f_y(X, A) - f_y(X-\epsilon_X, A-\epsilon_A)$. Therefore, the general unfaithfulness is then given by
\begin{align}
    \Delta(p) = &\mathbb{E}_{\epsilon_X, \epsilon_A} [\phi^\top W -  e^\top_v \delta(f)]^2
\end{align} To find the optimal $W^*$ that minimizes $\Delta(p)$, we take the derivative w.r.t $W$ and set it to zero (the integral in the expectation is over the random variables $\epsilon_X, \epsilon_A$ instead of $W$, therefore we can interchange the order of the derivative and the integral), which yields the following relation
\begin{align}
\mathbb{E}_{\epsilon_X, \epsilon_A}[\phi \phi^\top] W^* = \mathbb{E}_{\epsilon_X, \epsilon_A}[\phi \cdot e^\top_v \delta(f)]
\end{align} The solution exists as 
\begin{align}
    W^* = (\mathbb{E}_{\epsilon_X, \epsilon_A}[\phi \phi^\top])^{-1} \cdot \mathbb{E}_{\epsilon_X, \epsilon_A}[\phi\cdot e^\top_v \delta(f)]
\end{align} if the first part is invertible. The proof is generally similar to the case when $p$ is a linear function~\cite{NEURIPS2019_a7471fdc}.
\end{proof}

\section{Evaluation Details}
%\subsection{Dataset Statistics}
%\input{Tabel_dataset_statistics}

\subsection{Target GNN Models}\label{appendix:gcn-models}

We use a three-layer GNN for BA-shapes following the same architecture used in the public repository by~\citet{ying2019gnnexplainer}. We concatenate the output of each convectional layer and feed them into the final linear layer. This model archives 94\% test accuracy. For Cora and Citeseer, we use a 2-layer GNN without a dense layer. This model achieves 80.4\% test accuracy on Cora and 64.1\% on Citeseer. Training parameters are included in the code submission. 

\subsection{Implementation GNNExpl and PGExpl}\label{appendix:gnn-expl}
We use the implementation on Pytorch Geometric for GNNExpl\footnote{\url{https://pytorch-geometric.readthedocs.io/en/latest/\_modules/torch\_geometric/nn/models/gnn\_explainer.html\#GNNExplainer}}. We use all default parameters in this implementation except using \texttt{feat\_mask\_type='individual\_feature'} to get the feature masks and \texttt{return\_type='raw'} when our models provide logit outputs. \textbf{GNNExpl} drops edges with importance scores lower than 0.5 as the sigmoid function predicts whether an edge should be kept and we think 0.5 is a reasonable threshold. 

PGExpl was originally built in TensorFlow. We use the pytorch implementation\footnote{\url{https://github.com/LarsHoldijk/RE-ParameterizedExplainerForGraphNeuralNetworks}} acknowledged by \citet{luo2020parameterized} on their github page. We fixed a few bugs in dealing with edge cases in this implementation and reuse their hyper-parameters, e.g. number of epochs, temperatures. In BA-shapes, we find that training PGExpl with all points provides a slightly better results on ROC-AUC score so we train all nodes together. In Cora and Citeseer, we train each node individually. To create a subgraph explanation, we only keep top K\% edges. As \citet{luo2020parameterized} selects 100 for BAshapes, for nodes in which the average number of edges in its computation graph for a 3-layer GNN (our model) is around 739, we use $K = 100/739 = 13.5$.

\section{Other Experiments}

This section includes the details of several experiments mentioned in Sec.~\ref{sec:evaluations}. 

\subsection{Faithfulness Evaluations on Citeseer}\label{appendix:citeseer}

In this section, we include the results of measuring general unfaithfulness (Def.~\ref{def:general-unfaithfulness}) for baseline explanations and our proposed method, KEC, on Citeseer in addition to the results for BA-shapes and Cora shown in Table~\ref{tab:gf-individual}.

\begin{table*}[t]
  \small
  \centering
  \begin{tabular}{l|ccc}
  \toprule\toprule
  \textbf{Neighborhood}  & $\mathcal{U}(X, 0.2r)$ & $\mathcal{U}(A, 0.5)$ & $\mathcal{B}(A, 0.5)$ \\
  GNNExpl& 0.03&3.25  &0.60 \\
  GNNExpl (soft) &0.01 & 0.17 &0.21 \\
  PGExpl & - & 0.29 &0.48 \\
  SM & 1.40 $\times 10^{-3}$ & 3.70 $\times 10^{-4}$ & 0.23 \\
  IG (zero) & 2.16 $\times 10^{-3}$ & 0.10 & 0.43\\
  IG (random)& 0.02 & 1.90 $\times 10^{-3}$ & 0.26\\
  Linear & 0.04& 7.86 $\times 10^{-4}$ & 0.16\\
  KEC & \textbf{6.54} $\times \mathbf{10^{-4}}$ & \textbf{1.80} $\times \mathbf{10^{-4}}$ & \textbf{0.04} \\
  \bottomrule
  \end{tabular}
  \caption{General Unfaithfulness $\Delta(p)$ when node and edges are perturbed separately. Results are averaged over 200 nodes for Citeseer. Lower scores are better.} 
  \label{tab:gf-individual-citeseer}
\end{table*}

\begin{table}[t]
   \small
   \centering
   \begin{tabular}{l|cc|cc}
   \toprule\toprule
   & \multicolumn{2}{c}{\textbf{Cora}}&\multicolumn{2}{c}{\textbf{CiteSeer}} \\ 
   \midrule
   \textbf{Neighborhood} & $\mathcal{U}(X, 0.2r)$ & $\mathcal{U}(X, 0.5r)$ & $\mathcal{U}(X, 0.2r)$&  $\mathcal{U}(X, 0.5r)$ \\
   & $\times \mathcal{U}(A, 0.2)$ & $\times \mathcal{U}(A, 0.5)$ & $\times \mathcal{U}(A, 0.2)$&  $\times \mathcal{U}(A, 0.5)$ \\
   GNNExpl & 0.71 & 3.72  & 5.41  &2.30\\
   GNNExpl (soft)& 0.26 & 0.29  & 0.05  & 0.25\\
   SM & \textbf{0.02} & \textbf{0.26}  & \textbf{1.0}$\times \mathbf{10^{-3}}$ &   \textbf{0.05} \\
   IG (full)& 10.29 & 64.4  &13.71  & 85.69 \\
   Linear & 0.06& \textbf{0.26}  &0.17 & 0.20\\
   KEC & 0.10 & 0.42  & 0.026 &  0.15 \\
   \bottomrule
   \end{tabular} 
   \caption{General Unfaithfulness $\Delta(p)$ for mixed distributions where node and edge features are perturbed simultaneously. Results are averaged over 200 nodes and Lower scores are better.} 
   \label{tab:gf-mix}
\end{table}

\subsection{Mixed Distribution}\label{appendix:mixed-distribution}

We measure the general unfaithfulness $\Delta p$ for a mixed distribution by perturbing the node and edge features together. We denote this neighborhood as $\mathcal{U}(X, 0.5r) \times \mathcal{U}(A, 0.5)$ for example. This is a relatively larger neighborhood compared to all three distributions above. We do not perturb node features for BA-shapes as they are constants by construction. 

Notice that we increase the test samples size from 500 to 3000 per node in this case because this is a much larger neighborhood (details on how we get this number to follow in Sec.~\ref{sec:experiment-stability}). Since we are perturbing the node and edge features together, we use a baseline of zero vector for both node and edge features for IG, denoted as IG (full) in the table (we run test with a random baseline as well and it does make a big difference in this case). PGExpl is excluded because it does not attribute over node features. Results are shown in Table.~\ref{tab:gf-mix} (BAshapes is excluded because we are not interested in perturbing constant node features). 

\paragraph{Results.} We make the following observations: 1) The unfaithfulness scores for all methods are generally higher than the numbers they can get in $\mathcal{U}(X, \sigma r)$ or $\mathcal{U}(A, \sigma)$ with the difference on the order of magnitude. We hypothesize the reason for KEC to be less faithful in this distribution is because of the sample size, which we will discuss further in Appendix~\ref{appendix:approximation-error} as well. When solving KEC with a sufficient large number of samples is not feasible, we consider Saliency Map as a better choice in a mixed distribution in terms of the faithfulness; 3) Lastly, we again blame the unfaithfulness of IG for the choice of the baseline. Our results highlight the common baseline in image domain may not be a reasonable choice for graph data.

\subsection{Parameter Faithfulness}\label{appendix:paramter-faithfulness}
In Sec.~\ref{sec:evaluations}, we develop and evaluate the faithfulness of graph explanations on every single node but being faithful to the target model can also be considered in a larger scope -- that the explanations should be sensitive enough to the model's trainable parameters, i.e. a sanity check~\citep{10.5555/3327546.3327621}. We compare the performance of the explanations against the model's test accuracy with checkpoints saved during training. 

\subsubsection{Experiments and  Results}\label{appendix:auc-reproduction}

\begin{figure}[t]
\centering
    \begin{subfigure}[b]{0.45\textwidth}
        \centering
    \includegraphics[width=0.97\textwidth]{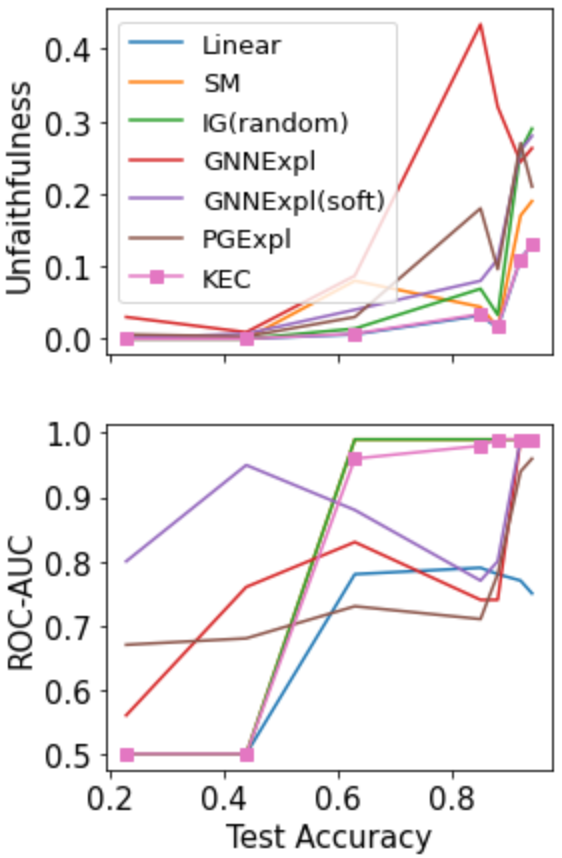}
        \subcaption{}\label{fig:auc}
    \end{subfigure}%
    \hfill
    \begin{subfigure}[b]{0.45\textwidth}
    \centering
    \includegraphics[width=\textwidth]{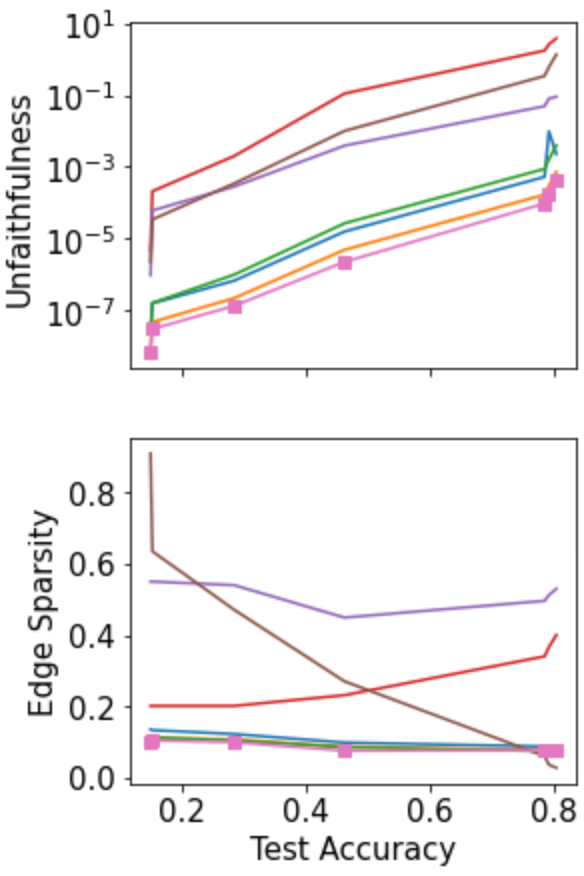}
      \subcaption{} \label{fig:sparsity}
    \end{subfigure}%
    \caption{(a) The performance of different explanations on BA-shapes (measured with general unfaithfulness and the ROC-AUC scores by comparing the highlighted edges with ground-truth edge labels) against the performance of the GNN model on the test nodes (measured with the classification accuracy score). (b) The performance of different explanations on Cora (measured with general unfaithfulness and the AUC score of edge sparsity) against the performance of the GNN model on the test nodes (measured with the classification accuracy score). }
\end{figure}

To quantitatively measure the performance of an explanation, we use ROC-AUC scores when a ground-truth label for the important edge is given, i.e. in BA-shapes, following the choice from the literature~\citep{ying2019gnnexplainer}. When a ground-truth is missing, i.e. Cora, we use Sparsity of important edges, i.e. $|A_t|/|A|$ where $A_t$ is an adjacency matrix with edges whose importance scores are all higher than $t$ and $A$ is the original adjacency matrix, as used in ~\citep{alsentzer2020subgraph}. Instead of choosing $t$, we firstly normalize edge importance scores to $[-1, 1]$ and measure the area of the curve when $t: 0\rightarrow 1$. We show the results for BA-shapes in Fig.~\ref{fig:auc} and for Cora in Fig.~\ref{fig:sparsity} for example.

Our findings are as follows: 1) the AUC scores and sparsity scores of SM, IG, KEC and Linear are better when the model is more accurate. The sparsity of PGExpl shows the highest sensitivity against the model's accuracy; 2) in both datasets all explanation methods become less faithful when the model becomes more accurate. We believe this is because when a model under-fits, it is more linear and its local behavior is easier to capture. An analogy to this is that gradient-based attributions all reduce to Saliency Map in a linear model~\citep{ancona2018towards}; 3) In Fig.~\ref{fig:auc}, we find that the ROC-AUC scores for SoftGNNExpl and PGExpl are pretty high even when the model is in its early stage of training while the correlations of the rest methods between explanation AUC and the test accuracy are stronger. We believe this demonstrates the necessity of refining the penalty functions $L$ (Def.~\ref{def: GNNExpl}) in the existing work to encourage the faithfulness.

% \begin{figure}[t]
%     \centering
%     \includegraphics[width=\textwidth]{Neurips-gnn/images/auc_horizongtal.png}
%     \caption{The performance of different explanations on BA-shapes (measured with general unfaithfulness and the ROC-AUC scores by comparing the highlighted edges with ground-truth edge labels) against the performance of the GNN model on the test nodes (measured with the classification accuracy score).}
%     \label{fig:auc}
% \end{figure}

\begin{figure*}[t]
    \begin{minipage}[b]{0.3\textwidth}
        \centering
        \includegraphics[width=0.96\textwidth]{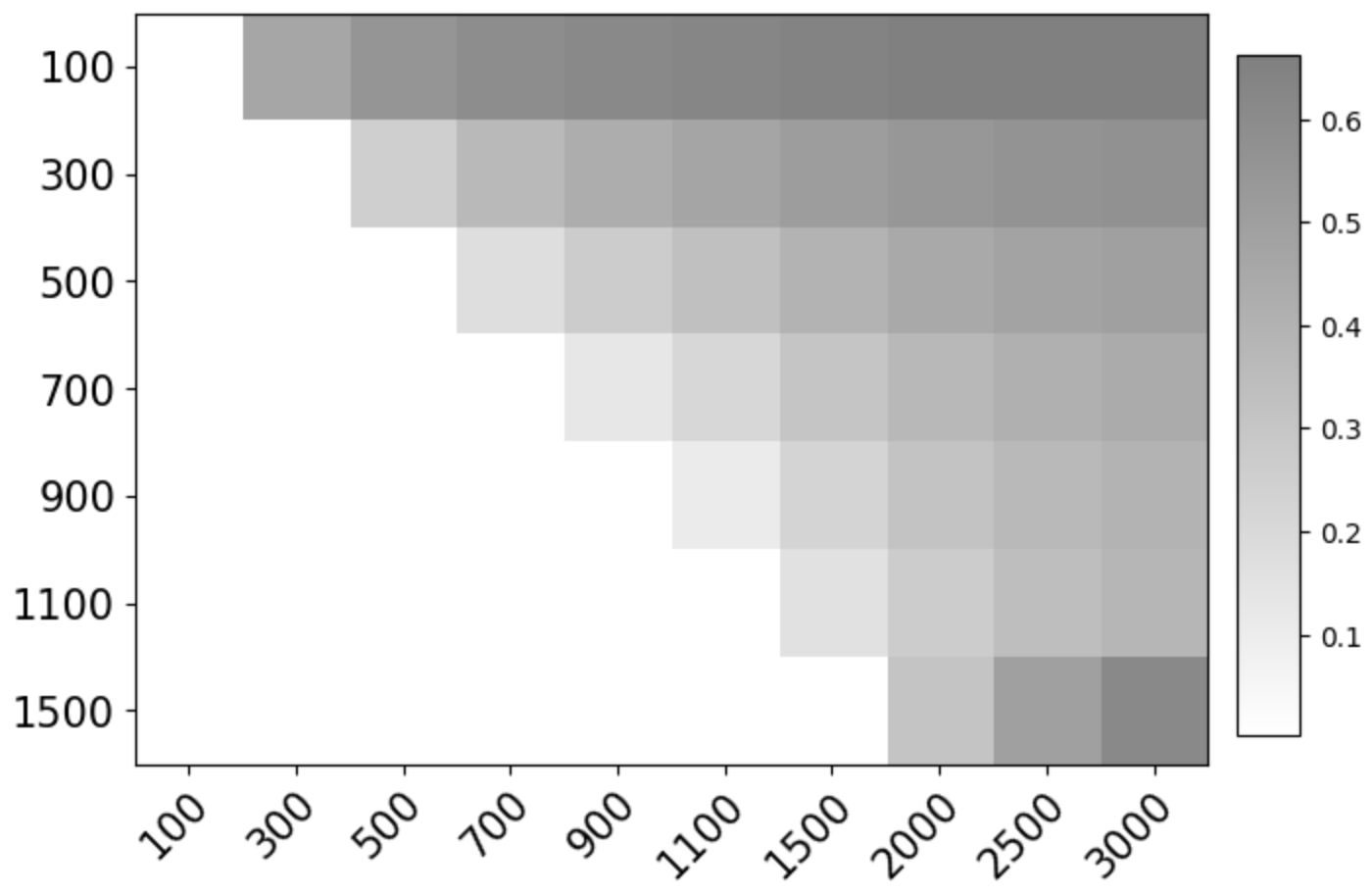}\\
        \subcaption{$\mathcal{U}(X, 0.5)$ }
    \end{minipage}%
    \hfill
    \begin{minipage}[b]{0.3\textwidth}
        \centering
        \includegraphics[width=\textwidth]{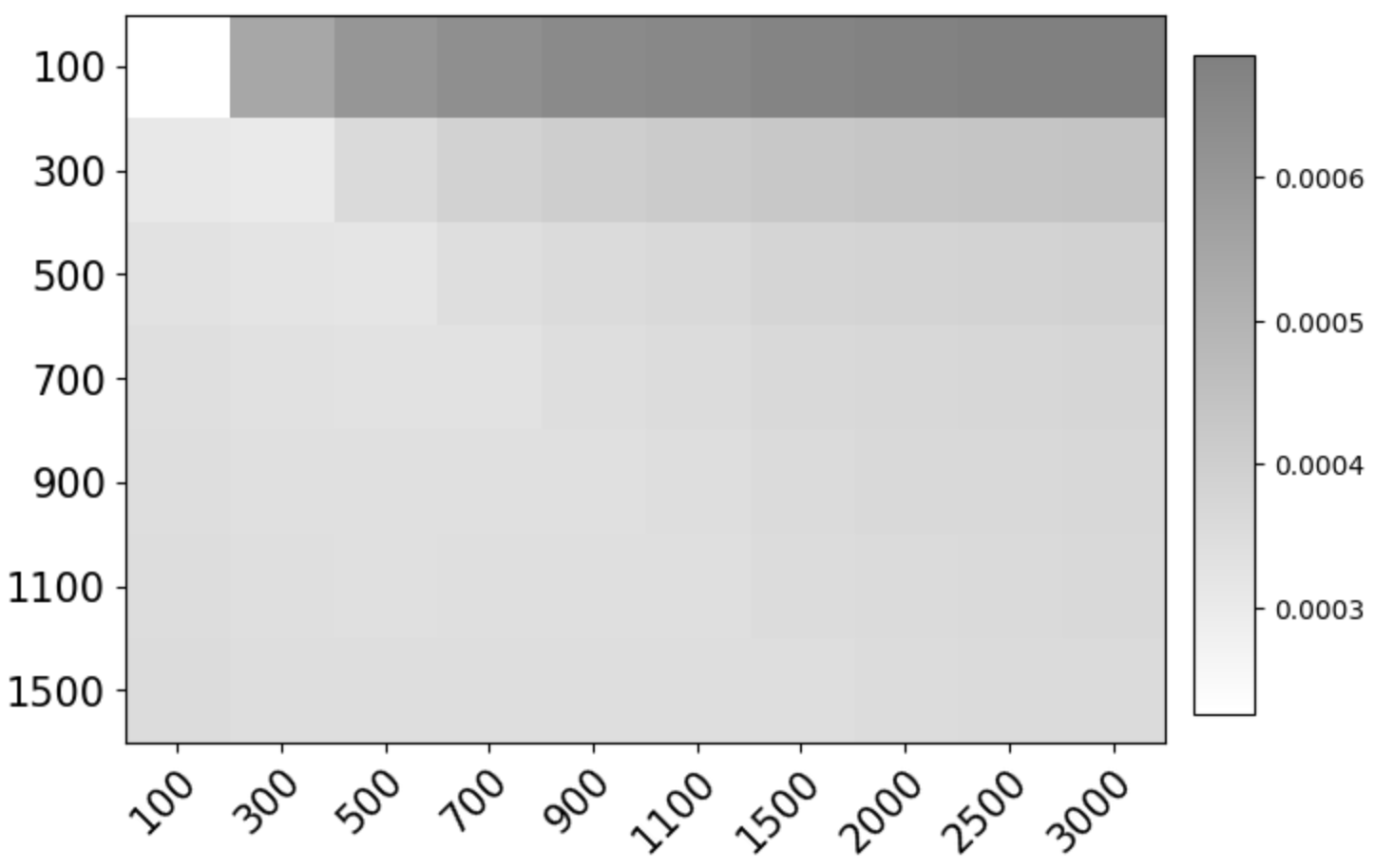}\\
        \subcaption{$\mathcal{U}(A, 0.5)$}
    \end{minipage}%
    \hfill
    \begin{minipage}[b]{0.3\textwidth}
        \centering
        \includegraphics[width=0.97\textwidth]{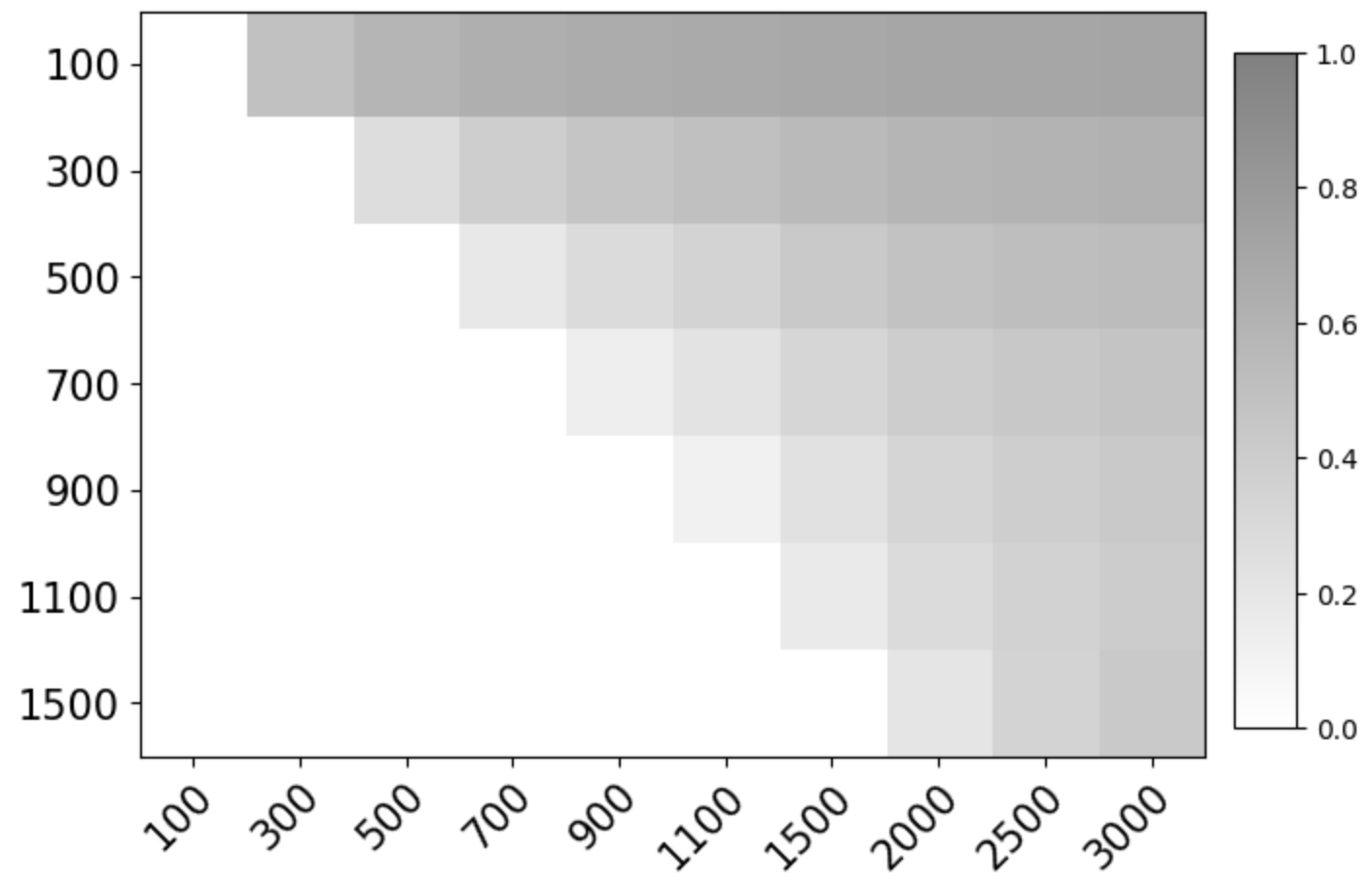}\\
        \subcaption{$\mathcal{U}(X, 0.5r) \times \mathcal{U}(A, 0.5)$}
    \end{minipage}%
   
\caption{General Unfaithfulness $\Delta(p)$ of KEC solved with truncated SVD using different number of samples (row) and tested with different number of samples (column) under different distribution of interests. Darker colors are less faithful and scales are shown in the color-bars by the side of each heatmap.}
 \label{fig:sample-size}
\end{figure*}

\subsubsection{A Note for Reproducing Scores from Related Work}\label{appendix:auc-reproduction}

\paragraph{ROC-AUC metric on GNNExpl.} Based on descriptions from \citet{ying2019gnnexplainer} and \citet{luo2020parameterized}, we use the attribution scores of each edge as the probability scores and compare this with the ground-truth labels of the edges. To do so, we first find k-hop neighborhood of the target node with \texttt{torch\_geometric.utils.subgraph} and call \texttt{sklearn.metrics.roc\_auc\_score()}. Because we are using newer Pytorch version (1.10.1+cu113 for Pytorch and 2.0.3 for Pytorch Geometric), we are not able to load the checkpoints from pretrained models. Instead, we retrain the model from scratch and report the ROC-AUC scores, which are 99\%, higher than the one, 92.5\%,  originally reported by \citet{ying2019gnnexplainer}. 

\begin{figure}[t]
    \centering
    \includegraphics[width=0.6\columnwidth]{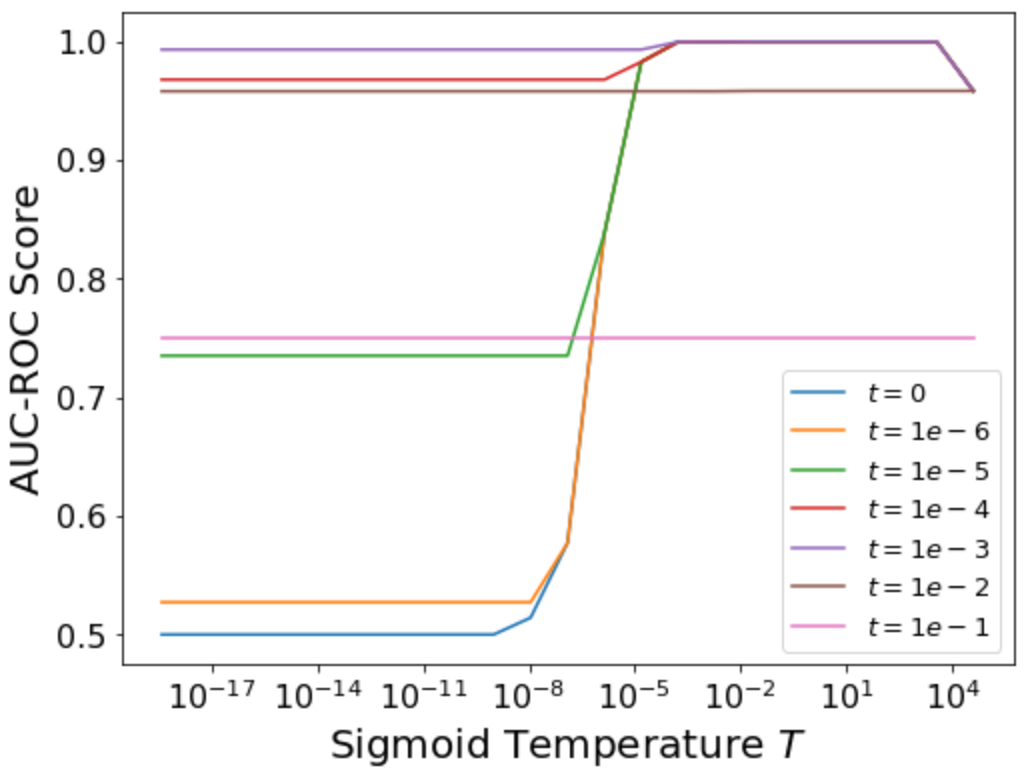}
    \caption{ROC-AUC scores change when the temperature $T$ of sigmoid function and the threshold $t$ vary when Saliency Map in BA-shapes}
    \label{fig:temperature}
\end{figure}

\paragraph{ROC-AUC metric on Gradient-based Approaches.} The AUC score for Saliency Map has been reported as 88.2\% by \citet{ying2019gnnexplainer} and reused by \citet{luo2020parameterized}.
We first describe the difference between our implementation of Saliency Map with \citet{ying2019gnnexplainer}. \citet{ying2019gnnexplainer} take the model's negative log probability\footnote{The minus sign is okay since \citet{ying2019gnnexplainer} later remove the sign of the gradient by taking its absolute value, which provides the same result as using the log probability. We refer to commit \texttt{c960a91} in their github repository.} w.r.t the input node and edge features and take the absolute value on the top of it; however, we directly take the derivative of the model's output logit on the target class w.r.t the input. Since all gradient-based approaches, i.e. Saliency Map, Integrated Gradient, KEC and Linear, return a vector with both positive and negative values, they are not in the range of [0, 1] and can be direclty usd as ``probabilities" as the attribution scores from GNNExpl and PGExpl. \citet{ying2019gnnexplainer} post-process the Saliency Map results by first taking the absolute values and feed it into a sigmoid function $s(x;T) = 1/[1+\exp(-x/T)]$, where $T$ is the sigmoid temperature and is set to 1 by \citet{ying2019gnnexplainer}, to map it into the range of [0, 1]. In practice, we find this mapping from the gradient space to the probability space is not well-justified because the signs of gradient values contain information of whether this features has positive or negative contributions. In practice, we find that by thresholding gradients, or explanations with both positive and negative values, and varying the temperature of sigmoid function, the ROC-AUC scores can span the entire possible range [0.5, 1.0] and an example of Saliency Map is shown in Fig.~\ref{fig:temperature}. We locate this issue and find it is caused by the numerical instability in computing the sigmoid function. As gradients can be as small as $1e-8$, when using \texttt{float} numbers, i.e. \texttt{np.float32}, and using temperature as $T=1$, tiny gradient values will overflow. To provide fair evaluations for Saliency Maps and all gradient-based methods, after the absolute function we firstly use a threshold to zero-out tiny gradients. The threshold is chosen as the minimal value that provides relative stable ROC-AUC scores over all temperature values from $10^{-18}$ to $10^{5}$. We pick $10^{-3}$ following our observations. Secondly, we run ROC-AUC scores with temperature sigmoid over the same temperature interval as mentioned above and use the average ROC-AUC scores for the explanation. This only applies to Saliency Map, Integrated Gradient, KEC, Linear in Sec.~\ref{fig:auc} and does not apply to GNNExpl and PGExpl as they are already in the range of [0, 1]. We believe this mapping from the gradient space to the probability space is not robust and should be replaced in the future.  

\subsection{Approximation Error}\label{appendix:approximation-error}

This section checks how many samples are sufficient to approximate the optimal parameters in KEC using truncated SVD and we show this in Fig.~\ref{fig:sample-size}. We use Cora as an example here. By varying the number of points we use to solve the parameter in KEC (numbers on the row axis) and the number of points we use to evaluate the faithfulness (numbers on the column axis), we get the heatmaps of unfaithfulness scores for three distributions $\mathcal{U}(X, 0.5)$, $\mathcal{U}(A, 0.5)$ and $\mathcal{U}(X, 0.5) \times  \mathcal{U}(X, 0.5)$ where darker grids are less faithful situations. We make the following observations: 1) in general we can minimize the unfaithfulness w.r.t a set of points equal or less than the samples we use to solve the parameter of KEC. This is observed from the diagonals of three heatmaps; 2) for perturbation only on edges, even with a small amount of samples, e.g. 100 samples, the unfaithfulness of KEC is still less than 0.001 even evaluated with 3000 samples; 3) for perturbations on node features and the mixed case, KEC may only be faithful for a subset of points used to solve the truncated SVD. We believe this is mainly because the neighborhood $\mathcal{U}(X, 0.5)$ is an $\ell_\infty$ ball with much more dimensions than that of $\mathcal{U}(A, 0.5)$. Namely, there are 1433 dimensions in the node features for Cora, which is about 3x larger than the average number, i.e. 414.8, of edges within a node's computation graph (a 3-hop subgraph in our case). Therefore, we use 3000 samples in Table~\ref{tab:gf-mix}.

\section{Anomaly Detection}\label{appendix:swat}
We follow the training strategy from \citet{deng2021swat-aaai}. Node features are one-dimensional time series data, which is the history of sensor readings over time. We use the physical structure of sensors as the adjacency information. During the training process, given the reading from $t$ to $t-4$, a GNN is trained to predict the reading at $t+1$ with MSE loss. During the test time, GNN's prediction of the sensor's reading is considered as \emph{normal values} and the actual observation is compared with the prediction. The maximum deviation between the model's deviation and the observation over node, when found higher than a threshold, will trigger the \emph{anomaly} flag and the node with the highest deviation is considered as the node under attach. The choice of threshold value is based on the model's prediction on a separated validation set. We refer the reader to  \citet{deng2021swat-aaai} for details and our code submission for the implementation.

\end{document}